\documentclass{article}

\usepackage{microtype}
\usepackage{graphicx}
\usepackage{subfigure}
\usepackage{booktabs} 

\usepackage{hyperref}

\usepackage[accepted]{icml2024}

\usepackage{amsmath}
\usepackage{amssymb}
\usepackage{mathtools}
\usepackage{amsthm}
\usepackage{wrapfig}
\usepackage{multirow}
\usepackage[capitalize,noabbrev]{cleveref}

\theoremstyle{plain}

\theoremstyle{definition}

\theoremstyle{remark}

\usepackage[textsize=tiny]{todonotes}
\usepackage[breakable, most]{tcolorbox}
\definecolor{boxcolor}{HTML}{2a8e1f}

\newtcolorbox{titlebox}[2][]{
    fonttitle=\itshape,
    title={#2},
    enhanced,
    colframe=boxcolor,
    coltitle=boxcolor,
    colback=white,
    left=0pt,
    right=0pt,
    boxed title style={opacityback=0,colframe=white,size=fbox,arc=0mm},
    attach boxed title to top left={yshift=-\tcboxedtitleheight/2,xshift=4mm}
}
\newtcolorbox{titlebox2}[2][]{
    fonttitle=\itshape,
    title={#2},
    width=9cm,
    enhanced,
    colframe=boxcolor,
    coltitle=boxcolor,
    colback=white,
    left=0pt,
    right=0pt,
    boxed title style={opacityback=0,colframe=white,size=fbox,arc=0mm},
    attach boxed title to top left={yshift=-\tcboxedtitleheight/2,xshift=4mm}
}

\begin{document}

\twocolumn[
\icmltitle{Improving Adversarial Energy-Based Model via Diffusion Process}



\icmlsetsymbol{equal}{*}

\begin{icmlauthorlist}
\icmlauthor{Cong Geng}{vivo}
\icmlauthor{Tian Han}{stevens}
\icmlauthor{Pengtao Jiang}{vivo}
\icmlauthor{Hao Zhang}{vivo}
\icmlauthor{Jinwei Chen}{vivo}
\icmlauthor{Søren Hauberg}{dtu}
\icmlauthor{Bo Li}{vivo}
\end{icmlauthorlist}

\icmlaffiliation{stevens}{Department of Computer Science, Stevens Institute of Technology, USA}
\icmlaffiliation{vivo}{vivo Mobile Communication Co., Ltd, China}
\icmlaffiliation{dtu}{Technical University of Denmark, Copenhagen, Denmark}

\icmlcorrespondingauthor{Søren Hauberg}{sohau@dtu.dk}
\icmlcorrespondingauthor{Bo Li}{ibra@vivo.com}

\icmlkeywords{Machine Learning, ICML}

\vskip 0.3in
]



\printAffiliationsAndNotice{}  

\begin{abstract}
Generative models have shown strong generation ability while efficient likelihood estimation is less explored. Energy-based models~(EBMs) define a flexible energy function to parameterize unnormalized densities efficiently but are notorious for being difficult to train.
Adversarial EBMs introduce a generator to form a minimax training game to avoid expensive MCMC sampling used in traditional EBMs, but a noticeable gap between adversarial EBMs and other strong generative models still exists. Inspired by diffusion-based models, 
we embedded EBMs into each denoising step to split a long-generated process into several smaller steps. Besides,  we employ a symmetric Jeffrey divergence and introduce a variational posterior distribution for the generator's training to address the main challenges that exist in adversarial EBMs. 
Our experiments show significant improvement in generation compared to existing adversarial EBMs, while also providing a useful energy function for efficient density estimation.
\end{abstract}

\section{Introduction}
\label{introduction}
Energy-based models~(EBMs) are known as one type of generative models that draw inspiration from physics and have been widely studied in machine learning~\cite{Hopfield2554,hinton1983optimal,10.5555/104279.104290}. EBMs define an unnormalized probability distribution over data space from a Gibbs density, which can be useful for several visual tasks, such as image classification~\cite{grathwohl2019your}, out-of-distribution~(OOD) detection~\cite{liu2020energy} or semi-supervised learning~\cite{gao2020flow}.  
However, training an EBM using maximum likelihood estimation can be challenging due to the lack of a closed-form expression for the normalization constant.
MCMC-based EBMs~\cite{du2019implicit,nijkamp2019learning} evaluate the gradient of the objective through Markov chain Monte Carlo~(MCMC) sampling on the defined energy function, which can be computationally expensive for both training and sampling. Adversarial EBMs~\cite{nomcmc,geng2021bounds} introduce a \emph{generator} to form a minimax game between alternative optimization of this generator and energy function, allowing for MCMC-free EBM training and fast sampling.

Although adversarial EBMs have great potential in distribution modeling, they still have some limitations that can be mainly attributed to three reasons. First, as is pointed out in~\citet{mescheder2018training} and~\citet{geng2021bounds}, minimax training can be unstable if two alternative optimization steps are not well balanced.
This instability poses a significant challenge in fitting the marginal energy distribution through adversarial training, particularly when dealing with large-scale, complex, and multi-modal data distributions. Secondly, most adversarial EBMs adopt KL divergence to optimize the generator. Since KL divergence is an asymmetric measure of divergence, relying solely on KL divergence may not be sufficient~\cite{arjovsky2017wasserstein}. Thirdly,
optimizing the generator requires computing an intractable entropy term,
leading to a trade-off between generation and density estimation~\cite{nomcmc}.  

To address these issues, we get inspiration from diffusion-based models~\cite{ho2020denoising} which specify a diffusion process to transform a data distribution by adding Gaussian noise with multiple steps until obtaining an approximate Gaussian distribution. A denoising diffusion process is learned by minimizing the KL divergence between true conditional denoising distribution $q\left(\mathbf{x}_{t-1} \mid \mathbf{x}_t\right)$ and a parameterized conditional distribution $p_\theta\left(\mathbf{x}_{t-1} \mid \mathbf{x}_t\right)$ at each noise step. Diffusion Recovery Likelihood~(DRL)~\cite{gao2020learning} and Denoising diffusion GAN~(DDGAN)~\cite{xiao2021tackling} combined EBM or GAN with diffusion process respectively. They trained a sequence of EBMs or GANs by matching true and modeled denoising distributions at each time step. The efficacy of these methods demonstrates that matching two denoising distributions is more tractable, as sampling from these conditional distributions is much easier than sampling from their marginal distributions. 
However, DDGAN lacks a useful density estimate for its discriminator and DRL has only one energy function, making sampling inefficient. These limitations are precisely where the strengths of EBM lie. Based on these arguments, we also incorporate adversarial EBMs into diffusion processes at each noise step. Thus for each EBM, the target distribution is a conditional distribution that is less multi-modal and easier to learn. Besides, with our defined generated denoising distribution through a generator, a symmetric Jeffrey divergence~\cite{jeffreys1946invariant} can be easily employed to remedy inadequate fitting, and a variational posterior distribution is introduced to compute the entropy term. Therefore, three main challenges in training adversarial EBMs can be overcome.
To our knowledge, it is the first time that adversarial EBM has been well integrated into the framework of diffusion models. 
In summary, the contributions of our paper are as follows:
\begin{itemize}
	\item We propose an MCMC-free training framework for EBMs to incorporate a sequence of adversarial EBMs into a denoising diffusion process. This framework avoids MCMC in both training and sampling.
	\item  We learn EBMs by optimizing conditional denoising distributions instead of marginal distributions to alleviate the training burden. We train the generator by minimizing a symmetric Jeffrey divergence to help distribution matching and introduce a variational posterior distribution to compute the entropy term. 
	\item We demonstrate that our model achieves significant improvements on sample quality compared to existing adversarial EBMs. We also verify the ability of our energy function in density estimation.
	
\end{itemize}

\section{Related Work}
There is a long history of EBMs in machine learning dating back to Hopfield networks~\cite{Hopfield2554} and Boltzmann machines~\cite{hinton1983optimal}. Recently, deep EBMs have grown in popularity, especially for image generation, and can broadly be classified into two categories.
MCMC-based methods simulate Markov chains during training~\cite{du2019implicit, pang2020learning, arbel2020generalized} or sampling~\cite{song2020sliced,song2019generative}. This can be expensive, slow, and hard to control. Cooperative learning methods~\cite{Xie_Lu_Gao_Zhu_Wu_2020, xie2021learning, xie2022tale, cui2023learning} jointly train a generator and an energy function for MCMC teaching, using the generator as a fast initializer for Langevin sampling to alleviate the MCMC burden. They, however, remain inefficient. \looseness=-1

The other category is adversarial EBMs~\cite{zhai2016generative,geng2021bounds,kumar2019maximum,zhao2016energy}, which introduce a 'generator' to form a minimax game between the energy function and the introduced generator. \citet{han2019divergence,han2020joint} and~\citet{kan2022bi} explained adversarial EBM from the view of triangle divergence and bi-level optimization problem respectively. 
Adversarial EBMs inherit the advantages of GANs and avoid MCMC sampling, but they also bring the risk of unstable training. 

The original diffusion-based model~\cite{ho2020denoising,sohl2015deep} learns a finite-time reversed diffusion from a forward process and gets strong performance on image generation, but the sampling process is slow. Several works~\cite{song2020denoising,lu2022dpm,bao2022analytic} improve training-free samplers and significantly improve the sampling speed going from 1000 steps sampling to only a few. DDGAN~\cite{xiao2021tackling} proposed to model each denoising step using a multimodal conditional GAN, tackling the generative learning trilemma of fast sampling while maintaining strong mode coverage and sample quality. However, its discriminator fails to provide a density estimate, which is crucial for many tasks~\cite{du2023reduce,zhang2022energy,du2022learning}.

Although some diffusion-based methods~\cite{kingma2021variational,song2021maximum,lu2022maximum} use score functions or variational lower bounds to estimate the density of the data distribution, these methods tend to be complicated and inefficient,  which is precisely where energy-based models excel. DRL~\cite{gao2020learning} and CDRL~\cite{zhu2023learning} tried to combine EBM with a diffusion-based model and got state-of-the-art generation performance on several image datasets. Still, they relied on MCMC sampling which remains inefficient.

\section{Denoising Diffusion Adversarial EBM}
\label{sec:Method}

\paragraph{Preliminaries}
Let $\mathbf{x}\sim q(\mathbf{x})$ be a training example from an underlying data distribution. An EBM is specified in terms of an energy function $E_\theta:\mathcal{X}\to \mathbb{R}$, that is parameterized by $\theta$, defines a probability distribution over $\mathcal{X}$ from the Gibbs distribution:
\begin{equation}
	p_\theta(\mathbf{x})=\frac{1}{Z_\theta} \exp \left(E_\theta(\mathbf{x})\right),
\end{equation}
where $Z_\theta$ is the normalization constant (or partition function). In principle, any density can be described in this way for a suitable choice of $E_\theta$. We learn the EBM through maximum likelihood estimation~(MLE), i.e.\@ we seek $\theta^*$ that maximizes the data log-likelihood:
\begin{equation}
	\begin{aligned}
	&\mathcal{L}(\theta^*):=\max_\theta\mathbb{E}_{\mathbf{x} \sim q(\mathbf{x})}\left[\log p_\theta(\mathbf{x})\right]\\&=\max_{\theta}\{\mathbb{E}_{\mathbf{x} \sim q(\mathbf{x})}\left[E_\theta(\mathbf{x})\right]-\log Z_\theta\}.
		\end{aligned}
\end{equation}
The fundamental challenge for training EBMs is the lack of a closed-form expression for the normalization constant. A common way to solve this is to approximate the gradient of $\mathcal{L}(\theta)$ directly and apply gradient-based optimization:
\begin{equation}
	\hspace{-2mm}\frac{\partial \mathcal{L}}{\partial \theta}=\mathbb{E}_{\mathbf{x} \sim q(\mathbf{x})}\left[\frac{\partial}{\partial \theta} E_\theta(\mathbf{x})\right]-\mathbb{E}_{\mathbf{x} \sim p_\theta(\mathbf{x})}\left[\frac{\partial}{\partial \theta} E_\theta(\mathbf{x})\right].
\end{equation}
Estimating the above equation needs MCMC sampling from the energy distribution $p_\theta(\mathbf{x})$, which is both unstable and expensive. Adversarial EBMs~\cite{nomcmc,geng2021bounds,han2019divergence,kan2022bi} avoid MCMC by introducing a variational distribution $p_\phi$ to approximate $p_\theta$.
Therefore, a minimax game can be formed to alternately optimize two adversarial steps:
\begin{enumerate}
\item $\min\limits_{p_\phi}D(p_\phi,p_\theta)$ with a certain proper divergence $D$.
\item $\max\limits_{E_\theta}\left[\mathbb{E}_{\mathbf{x} \sim q(\mathbf{x})}\left[E_\theta(\mathbf{x})\right]-\mathbb{E}_{\mathbf{x} \sim p_\phi(\mathbf{x})}\left[E_\theta(\mathbf{x})\right]\right]$.
\end{enumerate}
This two-step training strategy can be interpreted as a bi-level optimization problem~\cite{Liu_Gao_Zhang_Meng_Lin_2022}, where the minimization step is a lower-level~(LL) subproblem and the maximization step an upper-level~(UL) subproblem. \citet{geng2021bounds} demonstrated that if $p_\phi$ fails to be optimized, it runs the risk of maximizing an upper bound, causing a noticeable performance gap compared with mainstream generative models in complex and sparse datasets.

\subsection{Adversarial EBM with Denoising Diffusion Process}
To make adversarial EBMs scalable to complex distributions, we borrow the diffusion-based framework to split the generation process into multiple steps. For each step, we only need to learn a conditional distribution rather than a complex marginal distribution.
Similar to DDPM~\cite{ho2020denoising}, we also use a Markov chain for the diffusion process. Specifically, for the forward process, we gradually add noise to the real data $\mathbf{x}_0 \sim q(\mathbf{x}_0)$ in $T$ steps with pre-defined variance schedule $\beta_t$:
\begin{align}
	q\left(\mathbf{x}_{1: T} \mid \mathbf{x}_0\right)=\prod_{t \geq 1} q\left(\mathbf{x}_t \mid \mathbf{x}_{t-1}\right), 
	\\
	 q\left(\mathbf{x}_t \mid \mathbf{x}_{t-1}\right)=\mathcal{N}\left(\mathbf{x}_t ; \sqrt{1-\beta_t} \mathbf{x}_{t-1}, \beta_t \mathbf{I}\right). \nonumber
	 \label{forward process}
	\end{align}
Then we can get the denoising distribution $q\left(\mathbf{x}_{t-1} \mid \mathbf{x}_t\right)=\frac{q(\mathbf{x}_{t-1})q(\mathbf{x}_{t}|\mathbf{x}_{t-1})}{q(\mathbf{x}_{t})}$  as our target distribution to learn.
In diffusion-based models, the regular training objective is to minimize the KL divergence between the true denoising distribution $q\left(\mathbf{x}_{t-1} \mid \mathbf{x}_t\right)$ and a modeled one $p_\theta\left(\mathbf{x}_{t-1} \mid \mathbf{x}_t\right)$ for each $t$, i.e.
\begin{equation}
	\mathcal{L}=\!-\!\sum_{t \geq 1} \mathbb{E}_{q\left(\mathbf{x}_t\right)}\!\left[D_{\mathrm{KL}}\!\left(q\left(\mathbf{x}_{t-1} \!\mid\! \mathbf{x}_t\right) \| p_\theta\left(\mathbf{x}_{t-1} \!\mid\! \mathbf{x}_t\right)\right)\right] \!+\! C.
	\label{elbo}
\end{equation}
Inspired by DRL~\cite{gao2020learning}, 
we design a sequence of conditional EBM using an energy function conditioned on $t$ to define our modeled denoising distribution, 
\begin{equation}
	\begin{aligned}
	p_\theta\left(\mathbf{x}_{t-1} \mid \mathbf{x}_t\right)&=\frac{\exp(E_\theta(\mathbf{x}_{t-1},t-1))q(\mathbf{x}_{t}|\mathbf{x}_{t-1})}{\tilde{Z}_{\theta,t}(\mathbf{x}_t)}
 \end{aligned}
 \label{p theta}
\end{equation}
\begin{equation*}
\tilde{Z}_{\theta,t}(\mathbf{x}_t)=\int  \exp(E_\theta(\mathbf{x}_{t-1},t-1))q(\mathbf{x}_{t}|\mathbf{x}_{t-1})\mathrm{d}\mathbf{x}_{t-1} 
\end{equation*}
where $E_\theta(\boldsymbol{x}_t,t): \mathbb{R}^D \times \mathbb{R} \rightarrow \mathbb{R}$ is the energy function parameterized by $\theta$.
Thus $\mathcal{L}$ can be simplified and expressed by energy function:
\begin{equation}
\begin{aligned}
 \mathcal{L}
	&=\sum_{t \geq 1}\mathbb{E}_{ q(\mathbf{x}_{t-1},\mathbf{x}_{t})}\big[E_\theta(\mathbf{x}_{t-1},t-1)\\
	&+\log q(\mathbf{x}_{t}|\mathbf{x}_{t-1})-\log \tilde{Z}_{\theta,t}(\mathbf{x}_t)\big].
\end{aligned}
\label{mle}
\end{equation}
It is easy to see that when Eq.~\eqref{mle} is optimized, $p_\theta(\mathbf{x}_t)=\frac{\exp{\left(E_\theta(\mathbf{x}_t,t)\right)}}{Z_{\theta,t}}=q(\mathbf{x}_t)$~(see Appendix).

Since $\tilde{Z}_{\theta,t}$ is intractable, if we follow the common choice to compute the gradient of the objective: 
\begin{equation}
	\begin{aligned}
	\frac{\partial\mathcal{L}}{\partial \theta}&=\sum_{t \geq 1}\mathbb{E}_{q(\mathbf{x}_{t-1},\mathbf{x}_{t})}\left[\frac{\partial}{\partial \theta} E_\theta(\mathbf{x}_{t-1},t-1)\right]\\&
	-\mathbb{E}_{ q(\mathbf{x}_t) p_\theta(\mathbf{x}_{t-1}|\mathbf{x}_{t})}\left[\frac{\partial}{\partial \theta} E_\theta(\mathbf{x}_{t-1},t-1)\right],
		\end{aligned}
  \label{MCMC_t}
\end{equation}
the second term has to be estimated by MCMC-sampling from the distribution $q(\mathbf{x}_t) p_\theta(\mathbf{x}_{t-1}|\mathbf{x}_{t})$~\cite{gao2020learning}, which can be computational demanding and unstable during training.

Therefore, we adopt an adversarial EBM by introducing a variational conditional distribution $p_\phi\left(\mathbf{x}_{t-1} \mid \mathbf{x}_t\right)$ parameterized by $\phi$ to form a two-step minimax game:
\begin{titlebox}{DDAEBM minimax game}
\begin{enumerate}
    \item $\min\limits_{p_\phi}\sum_{t \geq 1}\mathbb{E}_{q\left(\mathbf{x}_t\right)} \left[D\left(p_\phi(\mathbf{x}_{t-1}|\mathbf{x}_{t}),p_\theta(\mathbf{x}_{t-1}|\mathbf{x}_{t})\right)\right]$
    \item 
    $\begin{aligned}[t]
       \max\limits_{E_\theta}\sum_{t \geq 1}\big[&\mathbb{E}_{q(\mathbf{x}_{t-1},\mathbf{x}_{t})}E_\theta(\mathbf{x}_{t-1},t-1) -\\
       &\mathbb{E}_{q(\mathbf{x}_{t})p_\phi(\mathbf{x}_{t-1}|\mathbf{x}_{t})}E_\theta(\mathbf{x}_{t-1},t-1)\big] 
    \end{aligned}$
\end{enumerate}
\end{titlebox}
This adversarial training strategy alternately optimizes $p_\phi$ and $E_\theta$, maintaining one as fixed while optimizing the other. This gives a tractable and MCMC-free approach to EBM training. We simply refer to our proposed method as \textbf{D}enoising \textbf{D}iffusion \textbf{A}dversarial \textbf{E}nergy-\textbf{B}ased \textbf{M}odel~(DDAEBM).

\subsection{Minimization w.r.t.\@ $p_\phi$}
First, we need to have a specific form of our introduced variational conditional distribution $p_\phi(\mathbf{x}_{t-1}|\mathbf{x}_{t})$. DDIM~\cite{song2020denoising} and DDPM~\cite{ho2020denoising} define $p_\phi(\mathbf{x}_{t-1}|\mathbf{x}_{t})$ as a Gaussian distribution to match the Gaussian posterior distribution $q\left(\mathbf{x}_{t-1} \mid \mathbf{x}_t, \mathbf{x}_0\right)$. However, as demonstrated in DRL, this normal distribution approximation is only accurate when $\beta_t$ is small, it may not be reasonable when there are few denoising steps, as the denoising distributions can be complex and multimodal.
Therefore, we define $p_\phi(\mathbf{x}_{t-1}|\mathbf{x}_{t})$ with a reparameterization trick,
\begin{align}
p_\phi&\left(\mathbf{x}_{t-1} \mid \mathbf{x}_t\right) := \int p_\phi\left(\mathbf{x}_0 \mid \mathbf{x}_t\right) q\left(\mathbf{x}_{t-1} \mid \mathbf{x}_t, \boldsymbol{x}_0\right) \mathrm{d} \mathbf{x}_0  \nonumber \\
	&=\int p(\mathbf{z}) q\left(\mathbf{x}_{t-1} \mid \mathbf{x}_t, \mathbf{x}_0=G_\phi\left(\mathbf{x}_t, \mathbf{z}, t\right)\right) \mathrm{d} \mathbf{z},
\label{denoising distribution}
\end{align}
where $p_\phi\left(\mathbf{x}_0 \mid \mathbf{x}_t\right)$ is an implicit distribution imposed by a neural network called \emph{the generator} $G_\phi\left(\mathbf{x}_t, \mathbf{z}, t\right): \mathbb{R}^D \times \mathbb{R}^d \times \mathbb{R} \rightarrow \mathbb{R}^D$. Here, $p(\mathbf{z})$ is a $d$-dimensional latent variable following a standard Gaussian distribution $\mathcal{N}(\mathbf{z} ; \mathbf{0}, \mathbf{I})$, and $q\left(\mathbf{x}_{t-1} \mid \mathbf{x}_t,\mathbf{x}_0\right)$ is the posterior distribution defined in forward diffusion process. It is worth noting that this definition is also explored by~\citet{xiao2021tackling}, which has been verified to be effective in fitting complex and multi-modal conditional distributions, although they don't leverage an EBM structure to characterize the data distribution.

The KL divergence is, by far, the most common choice of divergence $D$ for adversarial EBM training~\cite{geng2021bounds,nomcmc,zhai2016generative}. However, given that the KL divergence is an asymmetrical metric, it compels $p_\phi(\mathbf{x}_{t-1}|\mathbf{x}_{t})$ to chase the major modes of $p_\theta(\mathbf{x}_{t-1}|\mathbf{x}_{t})$. Relying solely on the KL divergence as our objective can, therefore, be insufficient for the generator to effectively capture the energy distribution~\cite{geng2021bounds}. Following~\citet{kan2022bi}, we choose a symmetric Jeffrey divergence, to integrate both KL divergence and reverse KL divergence into our objective:
\begin{titlebox}{Ideal objective}
\begin{align}
\min_{p_\phi}&\sum_{t \geq 1}\mathbb{E}_{q\left(\mathbf{x}_t\right)} \big[D_{\mathrm{KL}}(p_\phi(\mathbf{x}_{t-1}|\mathbf{x}_{t})\|p_\theta(\mathbf{x}_{t-1}|\mathbf{x}_{t})) \nonumber \\
		&+D_{\mathrm{KL}}(p_\theta(\mathbf{x}_{t-1}|\mathbf{x}_{t})\|p_\phi(\mathbf{x}_{t-1}|\mathbf{x}_{t}))\big].
	\label{min objective}
\end{align}
\end{titlebox}
We divide our ideal objective into two terms for separate handling:
\begin{align}
   L_1 &= \sum_{t \geq 1}\mathbb{E}_{q\left(\mathbf{x}_t\right)}D_{\mathrm{KL}}(p_\phi(\mathbf{x}_{t-1}|\mathbf{x}_{t})\|p_\theta(\mathbf{x}_{t-1}|\mathbf{x}_{t})) \label{first}, \\
    L_2 &=  \sum_{t \geq 1}\mathbb{E}_{q\left(\mathbf{x}_t\right)}D_{\mathrm{KL}}(p_\theta(\mathbf{x}_{t-1}|\mathbf{x}_{t})\|p_\phi(\mathbf{x}_{t-1}|\mathbf{x}_{t})). \label{second}
\end{align}

\paragraph{Bounding the first term of the ideal objective.}
To minimize our ideal objective, we first handle Eq.~\eqref{first}. It can be simplified by omitting some unrelated terms and obtaining:
\begin{equation}
\begin{aligned}
\sum_{t \geq 1}-H[p_\phi(\mathbf{x}_{t-1}|\mathbf{x}_{t})]-\mathbb{E}_{q(\mathbf{x}_t)p_\phi(\mathbf{x}_{t-1}|\mathbf{x}_{t})}\log p_\theta(\mathbf{x}_{t-1}|\mathbf{x}_{t})
\end{aligned}
\label{first term}
\end{equation}
Computing the entropy of the generated distribution is always a challenging task in EBMs. Several approaches~\cite{kumar2019maximum,nomcmc,geng2021bounds} have been proposed for efficient entropy approximation, but these are all either time or memory-demanding. We need to compute a conditional entropy:
\vspace{-1mm}
\begin{equation}
\begin{aligned}
&	H[p_\phi(\mathbf{x}_{t-1}|\mathbf{x}_{t})]=	H[p_\phi(\mathbf{x}_{t-1},\mathbf{z}|\mathbf{x}_{t})]-H[p_\phi(\mathbf{z}|\mathbf{x}_{t-1},\mathbf{x}_{t})] 
\end{aligned}
\end{equation}
where 
\vspace{-1mm}
\begin{equation}
  p_\phi(\mathbf{x}_{t-1},\mathbf{z}|\mathbf{x}_{t})=  p(\mathbf{z}) q\left(\mathbf{x}_{t-1} \mid \mathbf{x}_t, \mathbf{x}_0=G_\phi\left(\mathbf{x}_t, \mathbf{z}, t\right)\right).
\end{equation}
$H[p_\phi(\mathbf{z}|\mathbf{x}_{t-1},\mathbf{x}_{t})]$
is intractable while $H[p_\phi(\mathbf{x}_{t-1},\mathbf{z}|\mathbf{x}_{t})]$ is a constant that can be ignored~(see Appendix). Minimizing Eq.~\eqref{first term} can be problematic since we do not have access to $H[p_\phi(\mathbf{z}|\mathbf{x}_{t-1},\mathbf{x}_{t})]$, and we instead minimize a variational upper bound by introducing an approximate Gaussian posterior $q_\psi(\mathbf{z}|\mathbf{x}_{t-1},\mathbf{x}_{t})$. Here the mean and variance are represented through an encoder parameterized by $\psi$ with $\mathbf{x}_{t-1}$, $\mathbf{x}_t$ and $t$ as the input. We can easily obtain:
\begin{equation}
    H[p_\phi(\mathbf{z}|\mathbf{x}_{t-1},\mathbf{x}_{t})] \leq -\mathbb{E}_{q(\mathbf{x}_t)p_\phi(\mathbf{x}_{t-1},\mathbf{z}|\mathbf{x}_{t})} \log q_\psi(\mathbf{z}|\mathbf{x}_{t-1},\mathbf{x}_{t}).
    \label{entropy upper bound}
\end{equation}
Expressing $\log p_\theta(\mathbf{x}_{t-1}|\mathbf{x}_{t})$ with Eq.~\eqref{p theta} and applying Eq.~\eqref{entropy upper bound} to Eq.~\eqref{first term},  the first term Eq.~\eqref{first} can be minimized by its simplified variational upper bound: 
\begin{titlebox}{Upper bound of $L_1$}
\begin{align}
	\label{kl final}
		L_1 \leq \sum_{t \geq 1} \mathbb{E}_{q(\mathbf{x}_t)p_\phi(\mathbf{x}_{t-1},\mathbf{z}|\mathbf{x}_{t})}\!\Big[\!&-\!E_\theta(\mathbf{x}_{t-1},t\!-\!1)  \\
		-\log q(\mathbf{x}_{t}|\mathbf{x}_{t-1}) &-\!\log q_\psi(\mathbf{z}|\mathbf{x}_{t-1},\mathbf{x}_{t})\Big]		\nonumber
\end{align}
\end{titlebox}

\paragraph{Bounding the second term of the ideal objective.}
We now proceed with the minimization of our ideal objective, treating the second term Eq.~\eqref{second}. This term can also be bounded from above by utilizing an Evidence Lower Bound~(ELBO) associated with our introduced Gaussian posterior $q_\psi(\mathbf{z}|\mathbf{x}_{t-1},\mathbf{x}_{t})$, since 
\begin{equation}
	\begin{aligned}
		\log p_\phi&(\mathbf{x}_{t-1}|\mathbf{x}_{t}) \geq
  -D_{K L}\left(q_\psi(\mathbf{z}|\mathbf{x}_{t-1},\mathbf{x}_{t} ) \| p(\mathbf{z})\right)\\
		&+\mathbb{E}_{q_\psi(\mathbf{z}|\mathbf{x}_{t-1},\mathbf{x}_{t})}\log q(\mathbf{x}_{t-1}|\mathbf{x}_{t}, G_\phi\left(\mathbf{x}_t, \mathbf{z}, t\right))
	\end{aligned}
	\label{ELBO}
\end{equation}
We can get the upper bound of Eq.~\eqref{second} by applying Eq.~\eqref{ELBO}:
\begin{titlebox}{Upper bound of $L_2$}
\begin{align}
	\label{second term}
		&\hspace{0mm} L_2 \leq \sum_{t \geq 1} \mathbb{E}_{q(\mathbf{x}_t)p_\theta(\mathbf{x}_{t-1}|\mathbf{x}_{t})}D_{K L}\left(q_\psi(\mathbf{z}|\mathbf{x}_{t-1},\mathbf{x}_{t} ) \| p(\mathbf{z})\right) \raisetag{2mm} \\ 
 &\hspace{-1mm}-\! \mathbb{E}_{q(\mathbf{x}_t)p_\theta(\mathbf{x}_{t\!-\!1}|\mathbf{x}_{t}) q_\psi(\mathbf{z}|\mathbf{x}_{t\!-\!1},\mathbf{x}_{t})}
		\!\log q(\mathbf{x}_{t\!-\!1}|\mathbf{x}_{t}, G_\phi\!\left(\mathbf{x}_t, \mathbf{z}, t\right)) \nonumber
\end{align}
\end{titlebox}
A Monte Carlo approximation of Eq.~\eqref{second term} requires sampling from $q(\mathbf{x}_t)p_\theta(\mathbf{x}_{t-1}|\mathbf{x}_{t})$. As $p_\theta(\mathbf{x}_{t-1}|\mathbf{x}_{t})$ is designed to fit $q(\mathbf{x}_{t-1}|\mathbf{x}_{t})$, we use samples of $q(\mathbf{x}_{t-1},\mathbf{x}_{t})$ with an importance ratio to replace those of $q(\mathbf{x}_t)p_\theta(\mathbf{x}_{t-1}|\mathbf{x}_{t})$, as done in BiDVL~\cite{kan2022bi}:
\begin{equation}
    p_\theta(\mathbf{x}_{t-1}|\mathbf{x}_{t})=\lambda (t-1)q(\mathbf{x}_{t-1}|\mathbf{x}_{t})
\end{equation}
The importance ratio is designed inspired by ACT~\cite{kong2023act}:
\begin{equation}
    \lambda (t-1)=w\left(t^\prime(t)\right)^{\log _{\frac{1}{2}}\left(\frac{w_{\text{mid }}}{w}\right)},
    \label{weight elbo}
\end{equation}
where $w$ and $w_{\text {mid }}$ are the hyperparameters selected based on the datasets. $t'(\cdot)$ is a function denoting the time of $\mathbf{x}_t$ in the Variance Preserving (VP) SDE~\cite{song2020score}, as our diffusion process can be viewed as the discretization of the continuous-time VP SDE~(see Appendix).
The ultimate overall objective for implementation of the minimization step is the sum of Eq.~\eqref{kl final} and Eq.~\eqref{second term} w.r.t $p_\phi$ and $q_\psi$.

\subsection{Maximization step w.r.t.\@ $E_\theta$}
After minimizing step w.r.t $p_\phi$ and $q_\psi$, we approximately assume $p_\phi(\mathbf{x}_{t-1}|\mathbf{x}_{t})=p_\theta(\mathbf{x}_{t-1}|\mathbf{x}_{t})$. Next, we should optimize the energy function by maximizing:
\begin{equation}
	\begin{aligned}
 \sum_{t \geq 1}\big[\mathbb{E}_{q(\mathbf{x}_{t-1},\mathbf{x}_{t})}&E_\theta(\mathbf{x}_{t-1},t-1)
		\\ & \hspace{-5mm}-\mathbb{E}_{q(\mathbf{x}_{t})p_\phi(\mathbf{x}_{t-1}|\mathbf{x}_{t})}E_\theta(\mathbf{x}_{t-1},t-1)\big]
	\end{aligned}
	\label{maximizing step}
\end{equation}
Similar to most adversarial EBMs, we also find it helpful to stabilize training by adding a $\ell_2$-regularizer for the gradient of our energy:
\begin{equation}
\frac{\gamma}{2} \mathbb{E}_{ q\left(\mathbf{x}_{t-1} \right)}\left[\left\|\nabla_{\mathbf{x}_{t-1}} \left[E_\theta\left(\mathbf{x}_{t-1}, t-1\right)+\log q(\mathbf{x}_{t}|\mathbf{x}_{t-1})\right]\right\|^2\right]
\end{equation}
where $\gamma$ is the regularization coefficient. This regularization is a gradient penalty that alleviates training instability due to insufficient training at the minimization step~\cite{nomcmc,kumar2019maximum}. 

\section{Experiments}
We evaluate our DDAEBM in different scenarios across different data scales from 2-dimension toy datasets to large-scale image datasets. We test our energy function mainly on toy datasets and MNIST datasets which are easy to visualize and intuitive to measure. For large-scale datasets, we focus on image generation. We further perform some additional studies such as out-of-distribution~(OOD) detection and ablation studies to verify our model's superiority. We briefly introduce the network architecture design, while additional implementation details are presented in the Appendix. For the generator trained on large image datasets, we adopt the same modified NCSN++ architecture as DDGAN~\cite{xiao2021tackling} or DDPM++ with a slightly different structure as described in Score SDE~\cite{song2020score}, where diffused samples $\boldsymbol{x}_{t}$, time $t$ and latent variable $\mathbf{z}$ are the input of the network. For the energy function, we adopt traditional NCSN++ or DDPM++ architectures from Score SDE except that we remove the last scale-by-sigma operation and replace it with a negative $\ell_2$ norm between input $\boldsymbol{x}_{t}$ and the output of the Unet~\cite{du2023reduce}, i.e.
\begin{equation}
	E_\theta\left(\mathbf{x}_{t}, t\right)=-\frac{1}{2}\|\mathbf{x}_{t}-f_\theta\left(\mathbf{x}_{t}, t\right)\|^2.
 \label{energy_definition}
\end{equation}

\subsection{Performance on 2D synthetic data}
\begin{figure}[t]
	\footnotesize
	\centering
	\renewcommand{\tabcolsep}{1pt} \renewcommand{\arraystretch}{0.1}
	\begin{tabular}{ccccccc}
		\multirow{2}{*}[2.2em]{\includegraphics[width=0.16\linewidth]{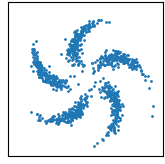}}&
		\includegraphics[width=0.16\linewidth]{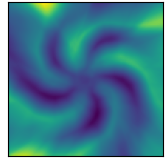} &
		\includegraphics[width=0.16\linewidth]{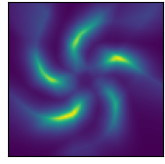} &
		\includegraphics[width=0.16\linewidth]{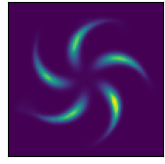} &
		\includegraphics[width=0.16\linewidth]{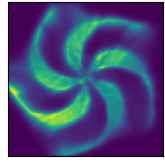}&
		\includegraphics[width=0.16\linewidth]{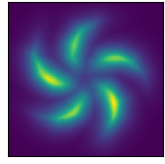} \\
		&\includegraphics[width=0.16\linewidth]{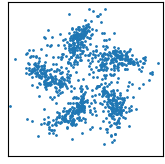} &
		\includegraphics[width=0.16\linewidth]{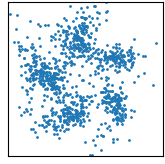} &
		\includegraphics[width=0.16\linewidth]{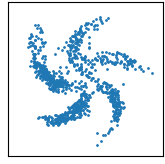} &
		\includegraphics[width=0.16\linewidth]{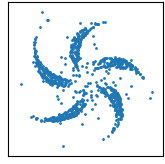}&
		\includegraphics[width=0.16\linewidth]{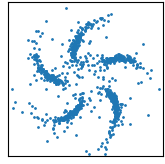}\\
		\multirow{2}{*}[2.2em]{\includegraphics[width=0.16\linewidth]{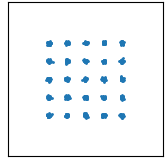}}&
		\includegraphics[width=0.16\linewidth]{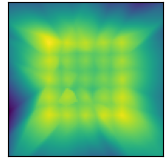} &
		\includegraphics[width=0.16\linewidth]{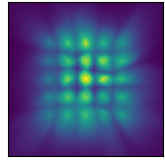} &
		\includegraphics[width=0.16\linewidth]{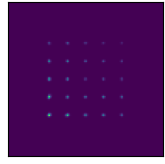} &
		\includegraphics[width=0.16\linewidth]{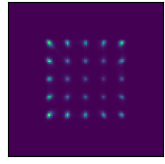}&
		\includegraphics[width=0.16\linewidth]{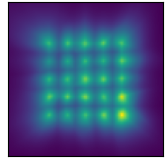}\\
		&\includegraphics[width=0.16\linewidth]{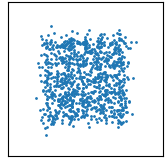} &
		\includegraphics[width=0.16\linewidth]{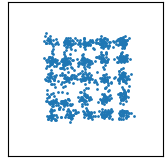} &
		\includegraphics[width=0.16\linewidth]{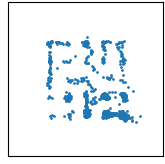} &
		\includegraphics[width=0.16\linewidth]{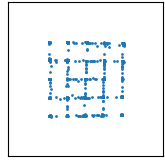}&
		\includegraphics[width=0.16\linewidth]{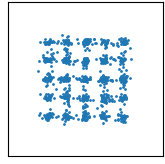}\\
  & MEG &  VERA & FCE &  BiDVL & DDAEBM 
	\end{tabular}
	\caption{Density estimation and generation on 25-Gaussians and pinwheel datasets. For each dataset, the first row shows the estimated densities and the second row shows generated samples.}
	\label{toy_exp} 
 \vspace{-10pt}
\end{figure} 
Fig.~\ref{toy_exp} demonstrates the density estimation and generation results of our DDAEBM on the 25-Gaussians and pinwheel datasets, which are two challenging toy datasets. We compare our model with several mainstream adversarial EBMs. These methods are all MCMC-free through training a generator and an energy function alternatively in two steps. We can observe that only our DDAEBM can get satisfying performance both for generation and density estimation on these two datasets. For pinwheel, most methods fail on density estimation except FCE and our DDAEBM. FCE~\cite{gao2020flow} trains an energy function using constrastive learning which has a strong ability on density estimation, but it uses a GLOW network as its generator which limits its ability on generation. For 25-Gaussians, although the generation of our DDAEBM is more dispersed than that of BiDVL around each mode, samples are mostly centered around each mode instead of being between two modes. 
\begin{table*}[t]
	\centering
	\caption{Log-likelihoods for NICE models as the energy function}
	\label{test LL}
	\resizebox{\linewidth}{!}{
		\begin{tabular}{c|cc|ccccccc}
			\toprule
			Method  & MLE  & MLE(t) & \begin{tabular}[c]{@{}c@{}}SSM\\ \cite{song2020sliced}\end{tabular}  & \begin{tabular}[c]{@{}c@{}}DSM\\ \cite{Vincent_2011}\end{tabular} & \begin{tabular}[c]{@{}c@{}}CoopNet\\ \cite{Xie_Lu_Gao_Zhu_Wu_2020}\end{tabular} & \begin{tabular}[c]{@{}c@{}}WGAN-0GP\\ \cite{mescheder2018training}\end{tabular} & \begin{tabular}[c]{@{}c@{}}MEG\\ \cite{kumar2019maximum} \end{tabular}  & \begin{tabular}[c]{@{}c@{}}VERA\\ \cite{nomcmc} \end{tabular}  & \begin{tabular}[c]{@{}c@{}}DDAEBM\\ (ours)\end{tabular} \\ \midrule
			Test LL & -791 & -879   & -2039 & -4363 & -1465   &   -1214       & -1023 & -1021 & $\textbf{-902}$    \\ \bottomrule
	\end{tabular}}
\end{table*}
\subsection{Fitting flow models with energy function}
Since our energy function is an unnormalized likelihood estimator with a difficult-to-estimate normalization constant, we consider an example where we can evaluate the exact log-likelihood associated with our energy function. Following \citet{nomcmc}, we train the NICE model~\cite{dinh2014nice} as an energy function on the MNIST dataset. NICE is a normalizing flow model that allows for exact likelihood computation and sampling. 
We get -879 on the MLE test using our network, which is close to the -791 of the traditional NICE model provided by \citet{song2020sliced}. Data is preprocessed following \citet{nomcmc}.
\begin{figure}[htbp]
	\footnotesize
	\centering
	\renewcommand{\tabcolsep}{1pt} \renewcommand{\arraystretch}{0.1}
	\begin{tabular}{cccc}
		     WGAN-0GP & MEG &VERA & DDAEBM \vspace{2pt}\\
	  	\includegraphics[width=0.25\linewidth]{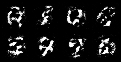} &
	  	\includegraphics[width=0.25\linewidth]{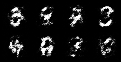} &
	  	\includegraphics[width=0.25\linewidth]{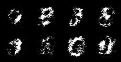} &
	  		\includegraphics[width=0.25\linewidth]{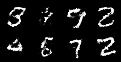} \vspace{2pt}\\
	  			\multicolumn{4}{c}{Exact Samples} \vspace{2pt}\\
		\includegraphics[width=0.25\linewidth]{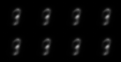}&
		\includegraphics[width=0.25\linewidth]{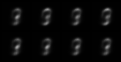}&
		\includegraphics[width=0.25\linewidth]{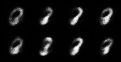}&
		\includegraphics[width=0.25\linewidth]{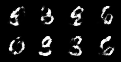}   \vspace{2pt}\\
		\multicolumn{4}{c}{Generator Samples} \\
	\end{tabular}
	\caption{Exact samples from the NICE model and generated samples from the generator for WGAN-based methods and our DDAEBM.}
	\label{mnist} 
\end{figure}
We observe that using the NICE network leads to worse performance of our DDAEBM than other WGAN-based methods such as WGAN-0GP, MEG, and VERA. The reason is that for WGAN-based methods, the gradients of the energy function's output are significantly smaller for fake samples compared to real samples, making the training of energy function similar to MLE. Therefore we add a weight of 0.1 for the second term in Eq.~\eqref{maximizing step} to reduce the effect of fake samples. We use the same trick on WGAN-based methods and find it can also improve their likelihood fitting but help little on generation even using a larger generator. We use an MLP network for our generator. Other experiment settings are the same as VERA~\cite{nomcmc}. Full experimental details can be found in the Appendix.

From Table~\ref{test LL} we observe that our method obtains the maximum log-likelihood on test data except MLE which maximizes the log-likelihood of training data as its objective. 
Fig.~\ref{mnist} shows the exact samples from NICE models and generated samples from the generator. Although all WGAN-based methods yield a good fit of log-likelihood, we can observe that WGAN-0GP and MEG fail to generate diverse and good-quality samples, VERA performs better but still has some mode collapse, our DDAEBM generates high-quality samples that match exact samples and reasonably capture the data distribution.
\subsection{Image generation}
\begin{figure*}[htbp]
	\footnotesize
	\centering
 \renewcommand{\arraystretch}{0.1}
 \setlength{\tabcolsep}{1pt} 
	\begin{tabular}{ccc}
			\includegraphics[width=0.3\linewidth]{ 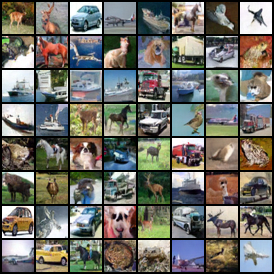} &
			 \includegraphics[width=0.3\linewidth]{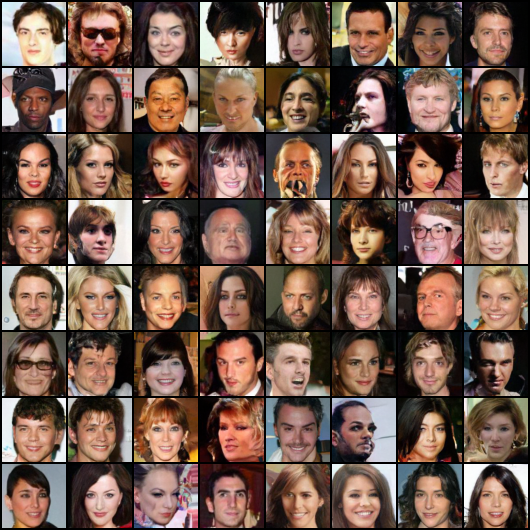}&
    \includegraphics[width=0.3\linewidth]{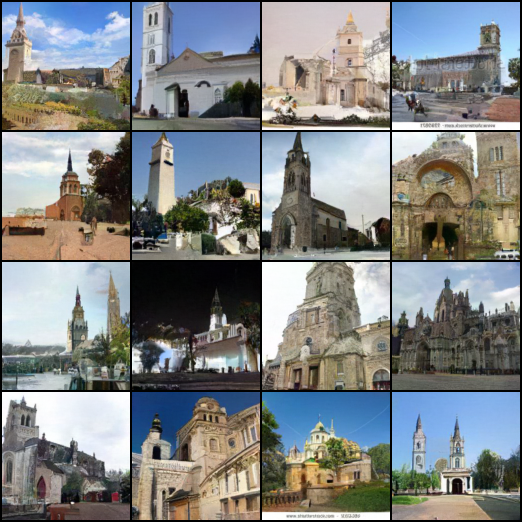}
		\end{tabular}
	\caption{Randomly generated images with DDAEBM on $32\times 32$ CIFAR-10, $64\times 64$ CelebA and $128\times 128$ LSUN church datasets.}
	\label{image_generation} 
\end{figure*}
To confirm the ability of our model to scale to large-scale datasets effectively, we conduct experiments of generation task on 32$\times$32 CIFAR-10~\cite{krizhevsky2009learning}, 64$\times$64 CelebA~\cite{liu2015deep}, and 128$\times$128 LSUN church~\cite{yu2015lsun} datasets. Each dataset represents a significant increase in complexity. All datasets are scaled to $[-1,1]$ for pre-processing. 
For quantitative results, we adopt commonly used Fréchet inception distance~(FID) and Inception Score~(IS) to evaluate sample fidelity and the number of function evaluations (NFE) to evaluate sampling time. Table~\ref{cifar10_quantitative} shows quantitative results on CIFAR-10. We observe our model achieves FID 4.82 and IS 8.86, outperforming existing adversarial EBMs by a significant margin and performing comparably to strong baselines from other generative models such as GANs and diffusion-based models, which struggle with efficient density estimation. 
\begin{table}[htbp]
	\centering
	\caption{Results for generation on CIFAR-10 dataset}
	\label{cifar10_quantitative}
	\renewcommand{\arraystretch}{1}
	\resizebox{\linewidth}{!}{%
		\begin{tabular}{ccccc}
			\toprule
			\multicolumn{1}{l}{}	&	Model                    & FID$\downarrow$   & IS$\uparrow$   & NFE$\downarrow$  \\ \midrule
			\multicolumn{5}{c}{Energy-based models}                   \\ \midrule
			\multirow{9}{*}{MCMC} &
			IGEBM~\cite{du2019implicit}        & 38.2  & 6.78 & 60   \\
			&JEM~\cite{grathwohl2019your}          & 38.4  & \textbf{8.76} & 20   \\
			& EBM-SR~\cite{nijkamp2019learning}       & 44.50 & 6.21 & 100  \\
			&	EBM-CD~\cite{du2020improved}       & 25.1  & 7.85 & 500  \\
			&			Hat EBM~\cite{hill2022learning}     & 19.30 & -    & 50   \\
            &       DAMC~\cite{yu2023learning}  & 57.72 & -    & 100 \\
			&	CoopNets~\cite{Xie_Lu_Gao_Zhu_Wu_2020}     & 33.61 & 6.55 & 50   \\
			&	VAEBM~\cite{xiao2020vaebm}        & 12.2  & 8.43 & 16   \\
        &	CLEL-large~\cite{lee2023guiding}   & \textbf{8.61}  & -    & 1200 \\
			&	DRL~\cite{gao2020learning} & 9.58  & 8.30 & 180 \\ 
			\midrule
			\multirow{7}{*}{No MCMC} &		EBM Triangle~\cite{han2019divergence}         & 28.96                & 7.30                 & 1                    \\
			&	MEG~\cite{kumar2019maximum}                & 35.02                & 6.49                 & 1                    \\
			&	VERA~\cite{nomcmc}                & 27.5                 & -                    & 1                    \\
			&	EBM-BB~\cite{geng2021bounds}               & 28.63                & 7.45                 & 1                    \\
			&	FCE~\cite{gao2020flow}                  & 37.30                & -                    & 1                    \\
			&	BiDVL~\cite{kan2022bi}                & 20.75                & -                    & 1                    \\
			&	\textbf{DDAEBM(ours)}          & \textbf{4.82}                 & \textbf{8.86}                 & 4       \\ \midrule
			\multicolumn{5}{c}{Other Generative Models}                   \\ \midrule
			&	SNGAN~\cite{miyato2018spectral}        & 21.7  & 8.22 & 1    \\
			&	BigGAN~\cite{brock2018large}                   & 14.73 & 9.22 & 1    \\
			&	StyleGAN2 w/ ADA~\cite{karras2020training}         & 2.92  & 9.83 & 1    \\ 
			&NCSN-v2~\cite{song2020improved}                  & 10.87 & 8.40 & 1000 \\
			&DDIM~\cite{song2020denoising}                     & 4.67  & 8.78 & 50   \\
			&Score SDE~\cite{song2020score}                & \textbf{2.20}  & \textbf{9.89} & 2000 \\
			& DDGAN~\cite{xiao2021tackling} & 3.75  & 9.63 & 4    \\
			&GLOW~\cite{kingma2018glow}                     & 48.9  & 3.92 & 1    \\
			\bottomrule
		\end{tabular}
	}
\end{table} 
Note that our DDAEBM is still less performant than DDGAN even though they use the same generator structure and parameterization of the generated denoising model. We further add an additional denoising step~\cite{song2020improved} using our energy function to refine the generated samples, we find the FID score can be improved to 3.73, which is on par with DDGAN. This denoising step can be approximately viewed as a one-step MCMC using our energy function. With this minor modification, we successfully bridge the gap between DDGAN and DDAEBM in terms of generation performance. 
Besides, the energy function in DDAEBM is tasked with providing a density estimate, not merely functioning as a discriminator. Therefore, the energy function in DDAEBM carries a heavier workload than the discriminator in DDGAN. 
In contrast, the discriminator in DDGAN merely distinguishes between real and fake sample pairs, primarily concentrating on the quality of the samples rather than on optimizing data likelihood. This explains  why the network of our energy function is stronger than the discriminator of DDGAN.

For CelebA, we only report FID scores in Table~\ref{celeba_quantitative} since the IS score is not widely reported. Our model gets the best performance among adversarial EBMs and outperforms score matching-based model NCSNv2 and VAE-based model NVAE. 
\begin{table}[htbp]
	\centering
	\caption{FID scores on CelebA 64$\times$ 64 dataset}
	\label{celeba_quantitative}
	\begin{tabular}{cc}
		\toprule
		Model              & FID$\downarrow$   \\ \midrule
		SNGAN~\cite{miyato2018spectral}              & 50.4  \\
		COCO-GAN~\cite{lin2019coco}           & \textbf{4.0}   \\ \midrule
		NVAE~\cite{vahdat2020nvae}               & 14.74 \\ \midrule
		VAEBM~\cite{xiao2020vaebm}              & 5.31  \\
		NCSNv2~\cite{song2020improved}             & 26.86 \\
		DDPM~\cite{ho2020denoising}               & \textbf{3.50}  \\
		DRL~\cite{gao2020learning}       & 5.98  \\ \midrule
		joint EBM Triangle~\cite{han2020joint} & 24.7  \\
		BiDVL~\cite{kan2022bi}              & 17.24 \\
		\textbf{DDAEBM~(ours)}              & \textbf{10.29} \\ \bottomrule
	\end{tabular}
	\vspace{-2mm}
\end{table}
For LSUN church, Fig.~\ref{image_generation} depicts high-fidelity synthesis sampled from the generator. We calculate FID on 50,000 samples using a PyTorch implementation from DDGAN and our model achieves 13.80 for this metric.
\subsection{Out of distribution detection}
The energy function of EBMs can be viewed as an unnormalized density function that assigns low values on out-of-distribution~(OOD) regions and high values on data regions, which is suitable for detecting OOD samples. To test this, we first train our DDAEBM on CIFAR-10 and calculate the energy $E_\theta(\mathbf{x}_{0},0)$ on in-distribution images from the CIFAR-10 test set and out-of-distribution images from datasets including SVHN~\cite{netzer2011reading}, Texture~\cite{cimpoi2014describing}, CIFAR-100~\cite{krizhevsky2009learning} and CelebA. Following~\citet{xiao2020vaebm}, we use the area under the ROC curve~(AUROC) as a quantitative metric on the energy scores, where high AUROC indicates that the model correctly assigns low density to OOD samples. Results are shown in Table~\ref{AUROC}, from where we can see our model achieves comparable performance with most of the baselines chosen from recent EBMs, VAEs, and Glow. Our model performs the best on SVHN and Texture datasets while on CIFAR-100 and CelebA it performs at a moderate level. Note that VAEBM performs slightly better than ours on most datasets, but it requires MCMC sampling to refine its generation, which can be inefficient.
As is pointed out by \citet{zhang2021understanding}, no method can guarantee better than random chance performance without assumptions on
which out-distributions are relevant.
Hence we see no clear winner on this task across all the datasets and we should take the results with a grain of salt.
\begin{figure*}[htbp]
\footnotesize
	\centering
	\renewcommand{\tabcolsep}{1pt} \renewcommand{\arraystretch}{0.1}
	\begin{tabular}{cccc}
			\includegraphics[width=0.25\linewidth]{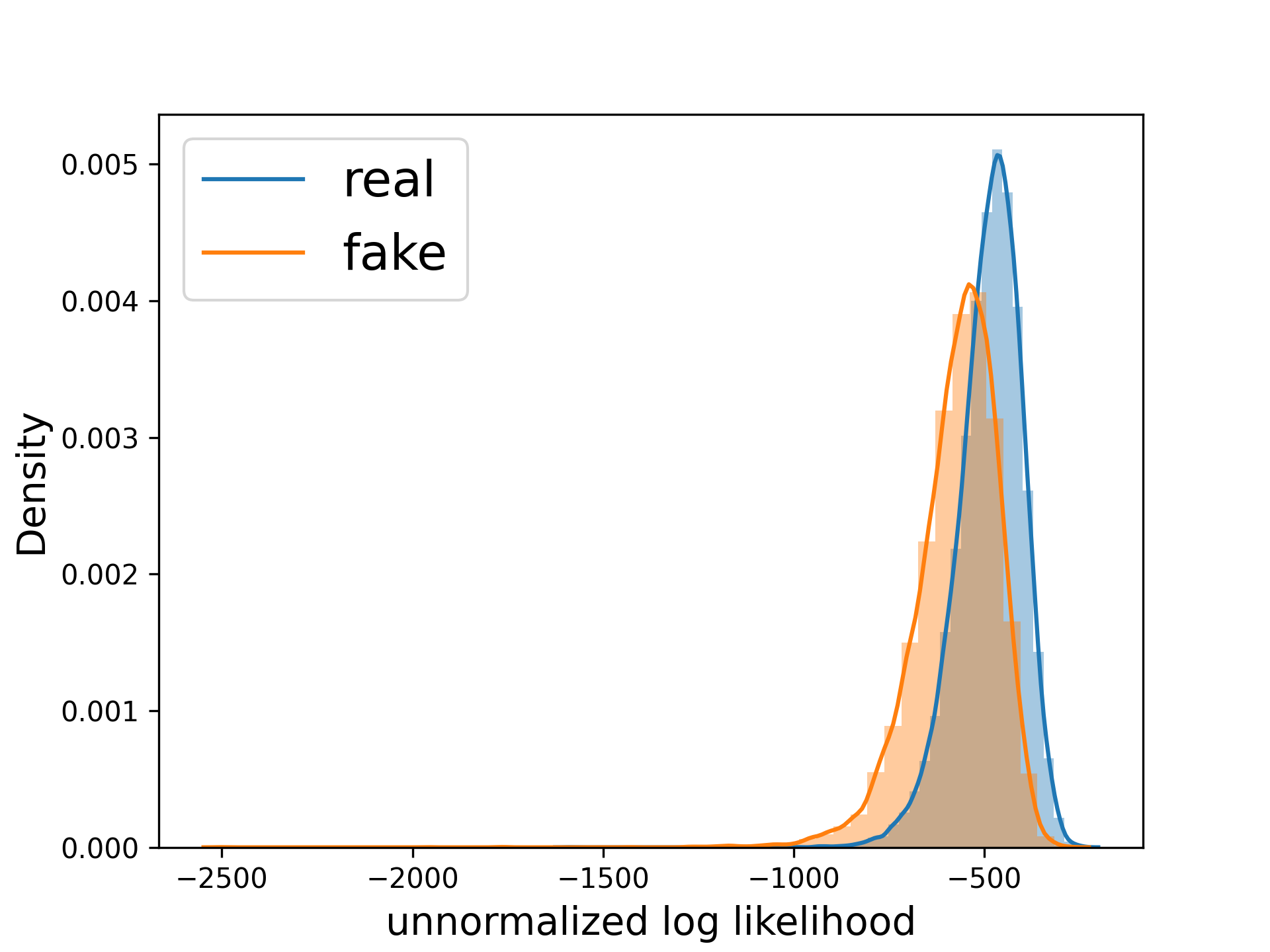} &
			\includegraphics[width=0.25\linewidth]{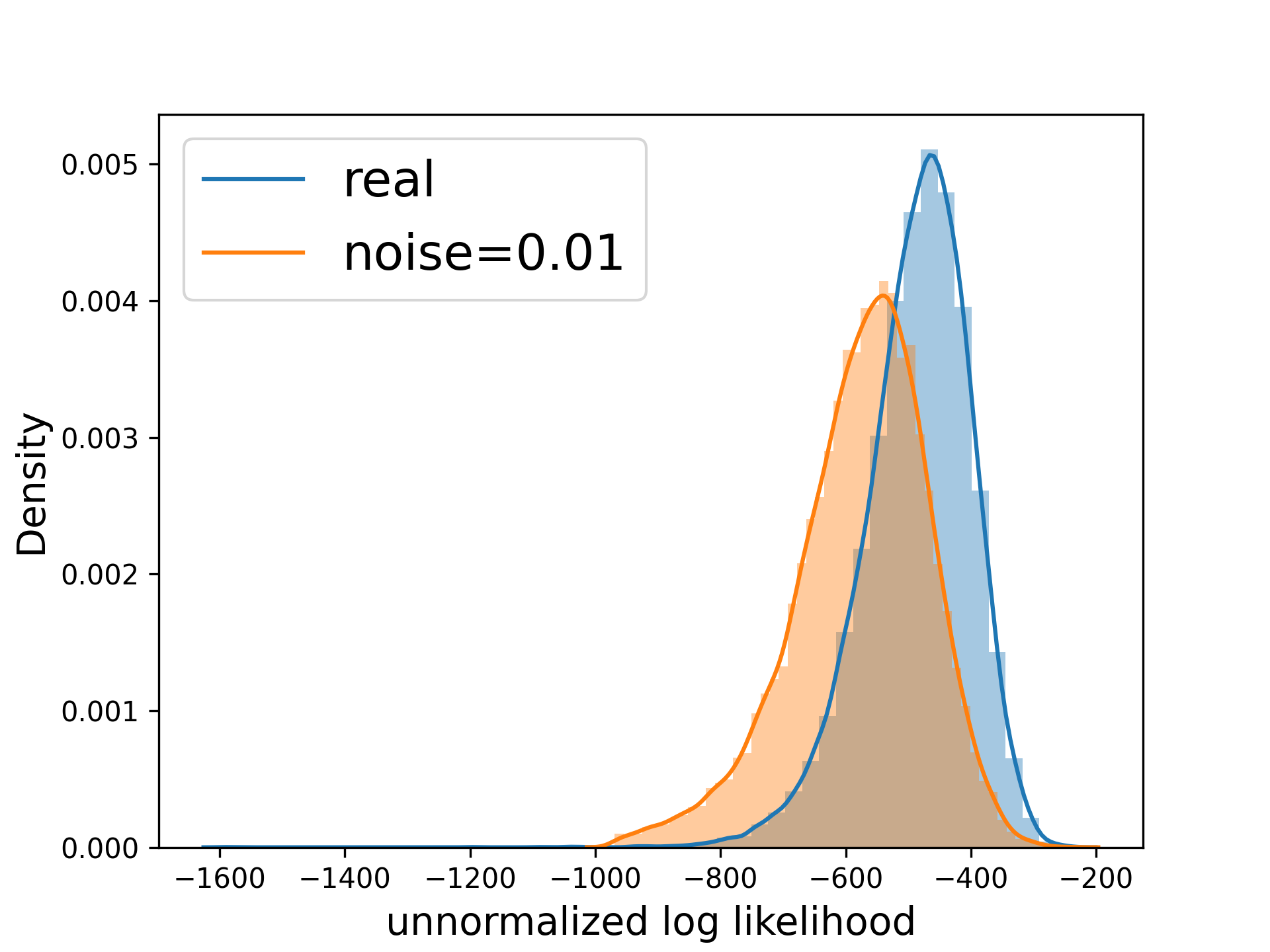} &
			\includegraphics[width=0.25\linewidth]{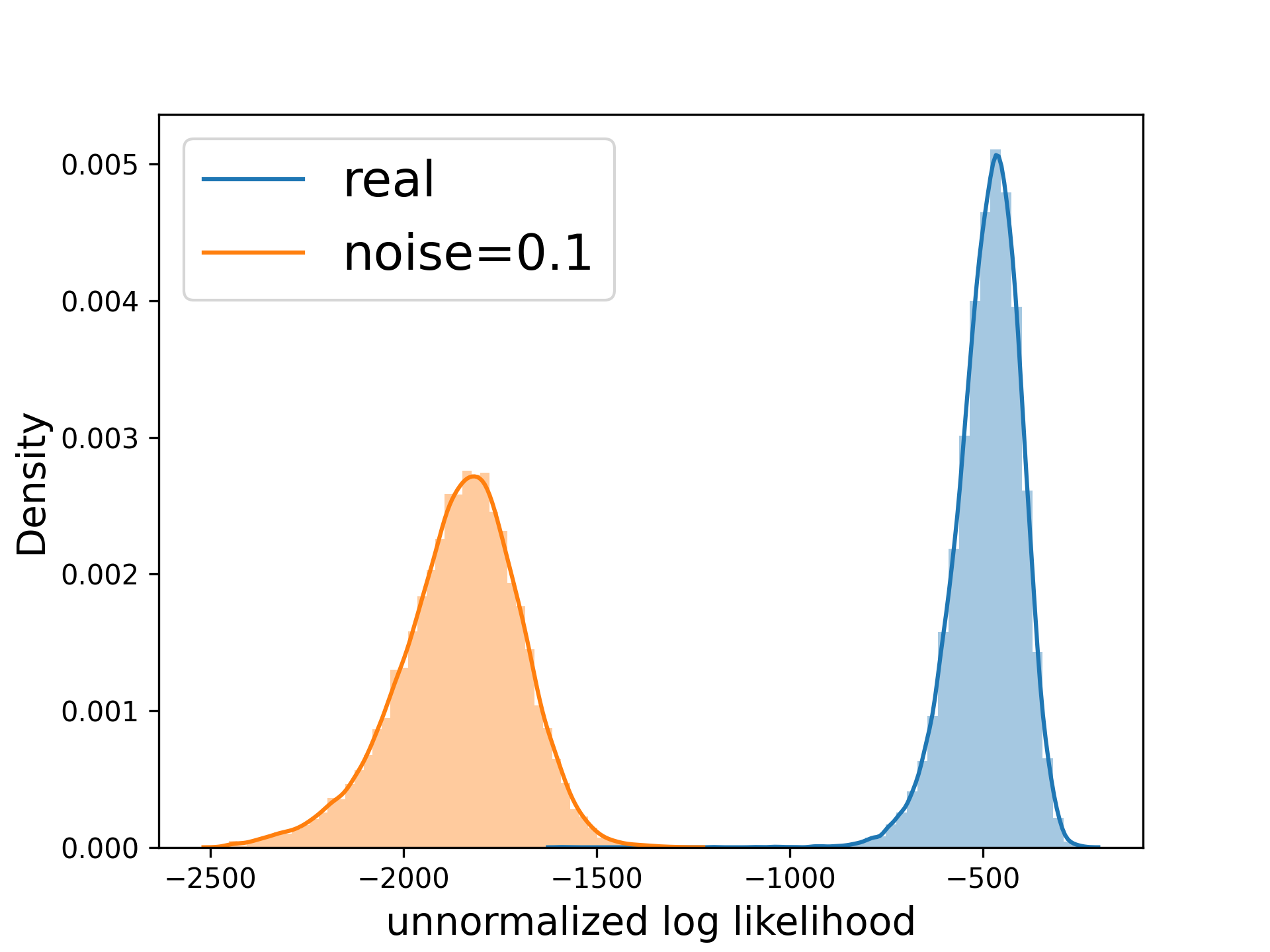} &
			\includegraphics[width=0.25\linewidth]{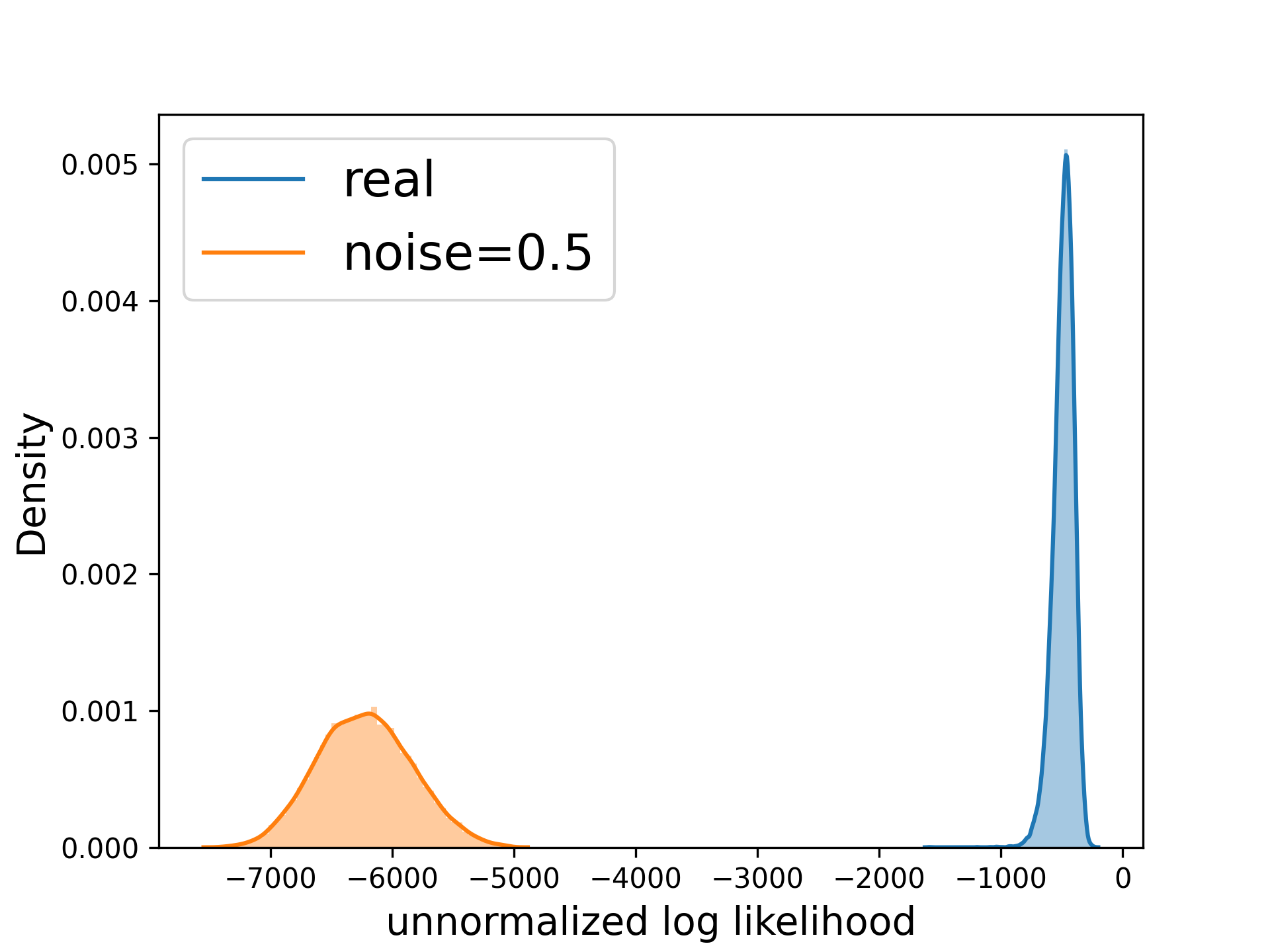} \\
		\end{tabular}
  \vspace{-3mm}
\caption{Histogram of unnormalized log-likelihood for comparison of real data and fake data or noisy data with the standard deviations being 0.01, 0.1, and 0.5. We provide the histogram comparison on CelebA test images. }
\label{energy_test} 
\end{figure*} 

Fig~\ref{energy_test} further shows histograms of the energy output at $t=0$ between the CelebA test dataset and several OOD distributions. For fake data, we diffuse real data $\mathbf{x}_0$ for one step and use $p_\phi(\mathbf{x}_0|\mathbf{x}_1)$ to generate fake data. For noisy data, we add noise to the real data with standard deviations of 0.01, 0.1, and 0.5. From Fig.~\ref{energy_test} we observe our energy function assigns high values to real data and low values to fake and noisy data. Additionally, the energy values gradually decrease as the added noise increases. Although fake data and noisy data with 0.01 standard deviation have imperceptible differences for visualization, their energy values still exhibit clear distinctions when compared to real data. This demonstrates our energy function's capability for OOD detection.
\begin{table}[htbp]
	\centering
\caption{AUROC for out-of-distribution detection on SVHN, Texture, CIFAR-100 and CelebA test datasets, with CIFAR-10 as the in-distribution dataset. }
\vspace{2pt}
\label{AUROC}
\resizebox{\linewidth}{!}{
\begin{tabular}{ccccc}
	\toprule
	Model               & SVHN                     & Texture & CIFAR-100      & CelebA        \\ \midrule
	IGEBM~\cite{du2019implicit}                & 0.63                     & 0.48    & 0.5           & 0.7           \\
	Glow~\cite{kingma2018glow}                & 0.05                     & -   & 0.55          & 0.57          \\
	NVAE~\cite{vahdat2020nvae}                & 0.42                     & -      & 0.56          & 0.68          \\
	SVAE~\cite{chen2018symmetric}                & 0.42                     & 0.5   & -             & 0.52          \\
	joint EBM Triangle~\cite{han2019divergence} & 0.68                     & 0.56    & -             & 0.56          \\
	VERA~\cite{nomcmc}                & \textbf{0.83}                     & -    & \textbf{0.73} & 0.33          \\
	BiDVL~\cite{kan2022bi}               & 0.76                     & -    & -             & \textbf{0.77} \\
	VAEBM~\cite{xiao2020vaebm}               & \textbf{0.83}                   & -      & 0.62          & \textbf{0.77} \\
	DDAEBM(ours)         & \textbf{0.83}            & \textbf{0.62}    & 0.60        & 0.70          \\ \bottomrule
\end{tabular}
}
\end{table}

\subsection{Ablation studies}
\paragraph{Importance of proposed modifications.} 
First, we investigate the effects of some proposed modifications in our model including the latent variable in our defined denoising distribution, introduced posterior $q_\psi(\mathbf{z}|\mathbf{x}_{t-1},\mathbf{x}_{t})$ and Jeffrey divergence for the generator's training. Table~\ref{ablation_modif} reports FID and IS scores on CIFAR-10 dataset and AUROC for the OOD task with CIFAR-10/SVHN as the in-distribution/out-of-distribution datasets. We replace sampling from $p(\mathbf{z})$ in Eq.~\eqref{denoising distribution} with all zero-vectors to remove the effect of $\mathbf{z}$ and observe the FID and IS scores are significantly worse. We also train our model with the most commonly used KL divergence and get similar performance on generation, but OOD performance significantly drops. We further
 remove $\log q_\psi(\mathbf{z}|\mathbf{x}_{t-1}, \mathbf{x}_{t})$-related term in our objective, which is equivalent to training a sequence of WGAN-0GPs embedded in a diffusion process.
 The generation and OOD performance are similar to the model trained with KL divergence. This experiment implies that asymmetric KL divergence and lack of $\log q_\psi(\mathbf{z}|\mathbf{x}_{t-1}, \mathbf{x}_{t})$ 
 lead to inaccurate learning of the energy function which, in turn, affects e.g. OOD performance. Symmetric Jeffrey divergence and entropy term $q_\psi(\mathbf{z}|\mathbf{x}_{t-1},\mathbf{x}_{t})$ ensure a better fit between the generated distribution and the energy function. This improves the energy training, resulting in improved OOD detection.
 More results can be found in the Appendix.
 \begin{figure}[htbp]
	\footnotesize
	\centering
		\includegraphics[width=0.8\linewidth]{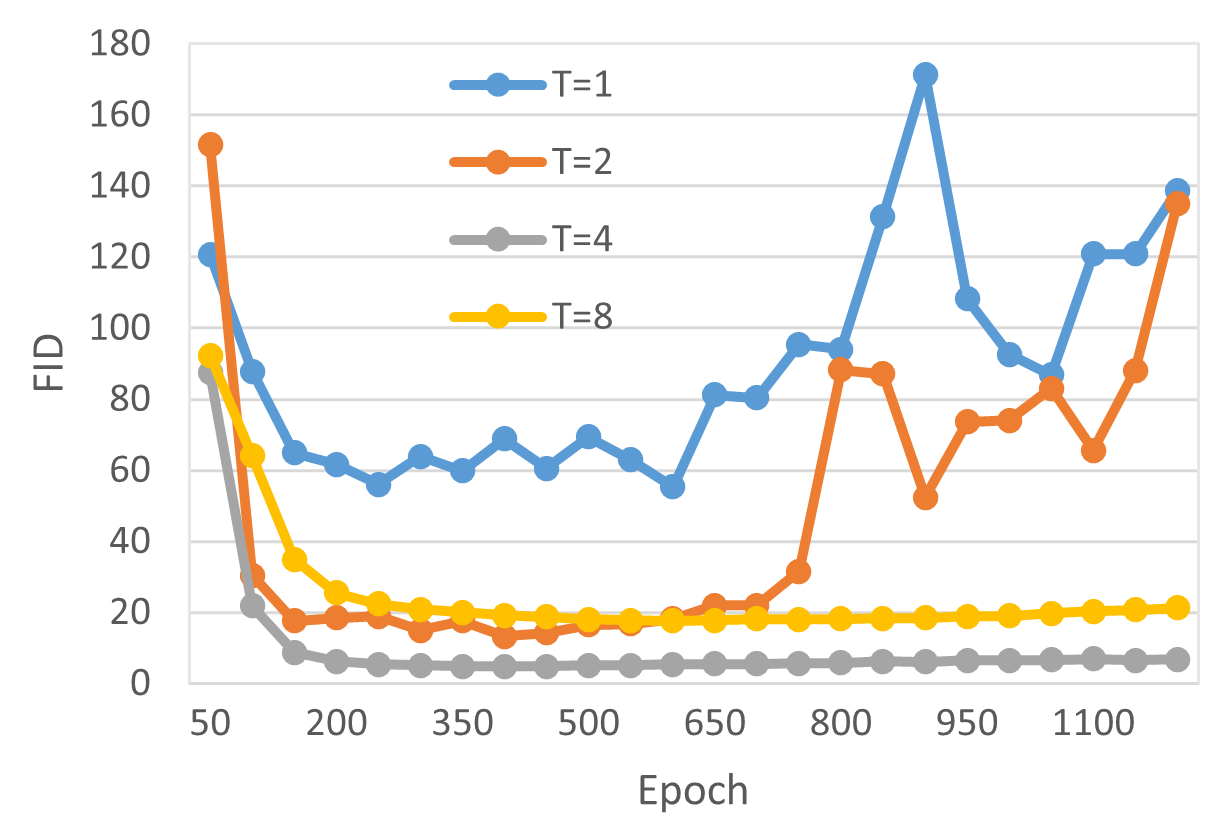} 
    \vspace{-3mm}
	\caption{The FID score vs.\@ the number of training epochs for a different number of time steps.}
	\label{abla_t} 
\end{figure}
\paragraph{Number of time steps.}  
We also examine the influence of varying the number of time steps $T$ by plotting the FID score versus training epochs for different time steps in Fig~\ref{abla_t}. Note that $T = 1$ corresponds to training a standard adversarial EBM, which leads to inadequate performance and unstable training. $T=2$ can get decent results very early but becomes unstable in later stages. $T=4$ gives excellent generation results and stable training which is consistent with DDGAN. When $T$ becomes larger, the training can be stable but there is an obvious degradation in performance. We hypothesize that although training for each step can be easier with a larger $T$, a higher capacity of the energy function is also required to accommodate more time steps. Thus, the choice of $T$ is important. We consistently chose $T=4$.
\begin{table}[htbp]
	\centering
	\caption{Ablation studies for some proposed modifications}
	\label{ablation_modif}
	\resizebox{\linewidth}{!}{
		\begin{tabular}{cccc}
			\toprule
			Model Variants     & FID$\downarrow$   & IS$\uparrow$  & \begin{tabular}[c]{@{}c@{}}OOD\\ (CIFAR-10/SVHN)$\uparrow$\end{tabular} \\ \midrule
			No latent variable & 10.09 & 8.25 & 0.23                                                          \\
			KL divergence            & 4.99  & \textbf{9.11}                  & 0.37                                                          \\
   KL w/o $q_\psi(\mathbf{z}|\mathbf{x}_{t-1},\mathbf{x}_{t})$         & 4.90  & 8.90                          & 0.38                                                          \\
			DDAEBM              & \textbf{4.82}  & 8.86    & \textbf{0.83}                                                          \\ \bottomrule
		\end{tabular}
	}
\end{table} 
\section{Conclusion}
We propose to integrate adversarial EBMs into the denoising diffusion process to learn the complex multimodal data distribution in several steps. This allows us to split a long generated process into several small steps, making each EBM easier to train. We define the generated denoising distribution by introducing a latent variable $\mathbf{z}$, which greatly accelerates sampling. With this definition, we employ a symmetric Jeffrey divergence to calibrate the training of our generator and introduce a variational posterior distribution $q_\psi(\mathbf{z}|\mathbf{x}_{t-1},\mathbf{x}_{t})$ to compute the entropy term, these operations address the long-standing training challenges existing in adversarial EBMs. Our model reduces the gap between adversarial EBMs and current mainstream generative models in terms of generation and provides a useful energy function with notable potential for a wide range of applications and downstream tasks.

\section*{Acknowledgements}
This work was supported by a research grant (42062) from VILLUM FONDEN. This project received
funding from the European Research Council (ERC) under the European Union’s Horizon 2020
research and innovation program (grant agreement 757360). The work was partly funded by the
Novo Nordisk Foundation through the Center for Basic Machine Learning Research in Life Science
(NNF20OC0062606).

\section*{Impact Statement}
This paper presents work whose goal is to advance the field of Machine Learning. There are many potential societal consequences of our work, as all high-capacity generative models carry the risk of being used for misinformation, and our model is no exception, none of which we feel must be specifically highlighted here.

\nocite{langley00}

\bibliography{example_paper}
\bibliographystyle{icml2024}

\newpage
\appendix
\onecolumn
\section{Extended derivations}
\label{sec:derivations}
\subsection{Derivations of Eq.~\eqref{mle}}
\begin{equation}
\begin{aligned}
    D_{\mathrm{KL}}\left(q\left(\mathbf{x}_{t-1} \mid \mathbf{x}_t\right) \| p_\theta\left(\mathbf{x}_{t-1} \mid \mathbf{x}_t\right)\right)=\int q\left(\mathbf{x}_{t-1} \mid \mathbf{x}_t\right)\log\frac{q\left(\mathbf{x}_{t-1} \mid \mathbf{x}_t\right)}{p_\theta\left(\mathbf{x}_{t-1} \mid \mathbf{x}_t\right)}d\mathbf{x}_{t-1}\\
    =\mathbb{E}_{q(\mathbf{x}_{t-1}|\mathbf{x}_{t})}\log q(\mathbf{x}_{t-1}|\mathbf{x}_{t})-\mathbb{E}_{q(\mathbf{x}_{t-1}|\mathbf{x}_{t})}\log p_\theta(\mathbf{x}_{t-1}|\mathbf{x}_{t})
\end{aligned}
\end{equation}
Thus the objective Eq.~\eqref{elbo} can be written as:
\begin{equation}
    \mathcal{L}=-\sum_{t \geq 1} \left[\mathbb{E}_{q(\mathbf{x}_{t-1}, \mathbf{x}_{t})}\log q(\mathbf{x}_{t-1}|\mathbf{x}_{t})-\mathbb{E}_{q(\mathbf{x}_{t-1}, \mathbf{x}_{t})}\log p_\theta(\mathbf{x}_{t-1}|\mathbf{x}_{t})\right]+C
\end{equation}
Since the first and the third term are independent of $\theta$, they can be disregarded during optimization. Thus the objective is only the second term. Plugging Eq.~\eqref{p theta} into the second term, we can obtain:
\begin{equation}
     \mathcal{L}=\sum_{t \geq 1} \mathbb{E}_{q(\mathbf{x}_{t-1}, \mathbf{x}_{t})}E_\theta(\mathbf{x}_{t-1},t-1)\\
 +\log q(\mathbf{x}_{t}|\mathbf{x}_{t-1})
		-\log \tilde{Z}_{\theta,t}(\mathbf{x}_t)
\end{equation}
\subsection{Derivations of the optimum in Eq.~\eqref{mle}}
When Eq.~\eqref{mle} gets optimum, due to the equivalence of Eq.~\eqref{elbo} and Eq.~\eqref{mle}, we can easily obtain $D_{\mathrm{KL}}\left(q\left(\mathbf{x}_{t-1} \mid \mathbf{x}_t\right) \| p_\theta\left(\mathbf{x}_{t-1} \mid \mathbf{x}_t\right)\right)=0$ for each time step, i.e.,
$q\left(\mathbf{x}_{t-1} \mid \mathbf{x}_t\right) = p_\theta\left(\mathbf{x}_{t-1} \mid \mathbf{x}_t\right)$,  which means:
\begin{equation}
	\begin{aligned}
	&E_\theta(\mathbf{x}_{t-1},t-1)+\log q(\mathbf{x}_{t}|\mathbf{x}_{t-1})
	-\log \tilde{Z}_{\theta,t}(\mathbf{x}_t)=\\
   &\log q(\mathbf{x}_{t-1})+\log q(\mathbf{x}_{t}|\mathbf{x}_{t-1})-\log q(\mathbf{x}_{t})
	\end{aligned}	
\end{equation}
Then we can obtain 
\begin{equation}
   E_\theta(\mathbf{x}_{t-1},t-1)-\log q(\mathbf{x}_{t-1})=
   \log \tilde{Z}_{\theta,t}(\mathbf{x}_t)-\log q(\mathbf{x}_{t})  
   \label{condition_eq} 
\end{equation}
The left side of the Eq.~\eqref{condition_eq} is just related to $\mathbf{x}_{t-1}$, while the right side is just related to $\mathbf{x}_t$, as $\mathbf{x}_{t-1}$ can respond to all $\mathbf{x}_t$ in the whole space of $\mathcal{X}$, then we can get $E_\theta(\mathbf{x}_{t-1},t-1)-\log q(\mathbf{x}_{t-1})=C$, therefore, for $\forall t$, we have 
\begin{equation}	\frac{\exp{\left(E_\theta(\mathbf{x}_t,t)\right)}}{Z_{\theta,t}}=q(\mathbf{x}_t),\\
	Z_{\theta,t}=\int \exp{\left(E_\theta(\mathbf{x}_t,t)\right)}d\mathbf{x}_t
\end{equation}
We already have  $p_\theta(\mathbf{x}_T)=q(\mathbf{x}_T)$ at $t=T$, 
when Eq.~\eqref{elbo} gets optimum, 
\begin{equation}
	\begin{aligned}
	&p_\theta(\mathbf{x}_{0: T})=q\left(\mathbf{x}_T\right) \prod_{t=1}^T p_\theta\left(\mathbf{x}_{t-1} \mid \mathbf{x}_t\right)\\
	&=q\left(\mathbf{x}_T\right) \prod_{t=1}^T q\left(\mathbf{x}_{t-1} \mid \mathbf{x}_t\right)=q(\mathbf{x}_{0: T})
\end{aligned}
\end{equation}
Therefore, we have $p_\theta(\mathbf{x}_t)=q(\mathbf{x}_t)=\frac{\exp{\left(E_\theta(\mathbf{x}_t,t)\right)}}{\boldsymbol{Z}_{\theta,t}}$.
\subsection{Derivations of Eq.~\eqref{MCMC_t}}
\begin{equation}
	\frac{\partial\mathcal{L}}{\partial \theta}=
	\sum_{t \geq 1}\mathbb{E}_{ q(\mathbf{x}_{t-1},\mathbf{x}_{t})}\big[\frac{\partial\left(E_\theta(\mathbf{x}_{t-1},t-1)
	-\log \tilde{Z}_{\theta,t}(\mathbf{x}_t)\right)}{\partial \theta}\big]
\end{equation}
where 
\begin{equation}
	\begin{aligned}
&\frac{\partial \log \tilde{Z}_{\theta,t}(\mathbf{x}_t)}{\partial \theta}=\frac{1}{\tilde{Z}_{\theta,t}}\frac{\partial\tilde{Z}_{\theta,t}}{\partial\theta}=\frac{1}{\tilde{Z}_{\theta,t}}\int  \frac{\exp(E_\theta(\mathbf{x}_{t-1},t-1))q(\mathbf{x}_{t}|\mathbf{x}_{t-1})}{\partial \theta}d\mathbf{x}_{t-1}\\
&=\int \frac{\exp(E_\theta(\mathbf{x}_{t-1},t-1))q(\mathbf{x}_{t}|\mathbf{x}_{t-1})}{\tilde{Z}_{\theta,t}}\frac{\partial E_\theta(\mathbf{x}_{t-1},t-1)}{\partial \theta}d\mathbf{x}_{t-1}\\
&=\int  p_\theta\left(\mathbf{x}_{t-1} \mid \mathbf{x}_t\right)\frac{\partial E_\theta(\mathbf{x}_{t-1},t-1)}{\partial \theta}d\mathbf{x}_{t-1}
\end{aligned}
\end{equation}
Therefore, we can obtain:
\begin{equation}
	\begin{aligned}
	\frac{\partial\mathcal{L}}{\partial \theta}=\sum_{t \geq 1}\mathbb{E}_{ q(\mathbf{x}_{t-1},\mathbf{x}_{t})}\frac{\partial E_\theta(\mathbf{x}_{t-1},t-1)}{\partial \theta}-\mathbb{E}_{q(\mathbf{x}_t) p_\theta(\mathbf{x}_{t-1}|\mathbf{x}_{t})}\frac{\partial E_\theta(\mathbf{x}_{t-1},t-1)}{\partial \theta}
		\end{aligned}
		\label{MCMC_correct}
\end{equation}

\subsection{Derivations of Eq.~\eqref{first term}}
\begin{equation}
    D_{\mathrm{KL}}(p_\phi(\mathbf{x}_{t-1}|\mathbf{x}_{t})\|p_\theta(\mathbf{x}_{t-1}|\mathbf{x}_{t}))
    =\mathbb{E}_{p_\phi(\mathbf{x}_{t-1}|\mathbf{x}_{t})}\log p_\phi(\mathbf{x}_{t-1}|\mathbf{x}_{t})-\mathbb{E}_{p_\phi(\mathbf{x}_{t-1}|\mathbf{x}_{t})}\log p_\theta(\mathbf{x}_{t-1}|\mathbf{x}_{t})
\end{equation}
Thus 
\begin{equation}
\begin{aligned}
    &\sum_{t \geq 1}\mathbb{E}_{q\left(\mathbf{x}_t\right)}D_{\mathrm{KL}}(p_\phi(\mathbf{x}_{t-1}|\mathbf{x}_{t})\|p_\theta(\mathbf{x}_{t-1}|\mathbf{x}_{t})) \\&=
    \sum_{t \geq 1}\mathbb{E}_{q(\mathbf{x}_t)p_\phi(\mathbf{x}_{t-1}|\mathbf{x}_{t})}\log p_\phi(\mathbf{x}_{t-1}|\mathbf{x}_{t})-\mathbb{E}_{q(\mathbf{x}_t)p_\phi(\mathbf{x}_{t-1}|\mathbf{x}_{t})}\log p_\theta(\mathbf{x}_{t-1}|\mathbf{x}_{t})\\
    &=\sum_{t \geq 1}-H[p_\phi(\mathbf{x}_{t-1}|\mathbf{x}_{t})]-\mathbb{E}_{q(\mathbf{x}_t)p_\phi(\mathbf{x}_{t-1}|\mathbf{x}_{t})}\log p_\theta(\mathbf{x}_{t-1}|\mathbf{x}_{t})
    \end{aligned}
\end{equation}
\subsection{Derivations of the entropy}
Since 
\begin{equation}
p_\phi(\mathbf{x}_{t-1},\mathbf{z}|\mathbf{x}_{t})=p(\mathbf{z}) q\left(\mathbf{x}_{t-1} \mid \mathbf{x}_t, \mathbf{x}_0=G_\phi\left(\mathbf{x}_t, \mathbf{z}, t\right)\right),
\end{equation}
according to the property of entropy~\cite{ORLITSKY2003751}
\begin{equation}
	H[\mathbf{x},\mathbf{z}]=H[\mathbf{z}]+H[\mathbf{x}|\mathbf{z}],
\end{equation}
we can get
\begin{equation}
H[p_\phi(\mathbf{x}_{t-1},\mathbf{z}|\mathbf{x}_{t})]=	H[\mathbf{z}]+H[q(\mathbf{x}_{t-1}|\mathbf{x}_{t},G_\phi\left(\mathbf{x}_t, \mathbf{z}, t\right))]
\end{equation}
Since  $p(\mathbf{z})$ and $q(\mathbf{x}_{t-1}|\mathbf{x}_{t},G_\phi\left(\mathbf{x}_t, \mathbf{z}, t\right))$ are both Gaussian distributions, 
according to~\cite{misra2005estimation}, a m-dimensional Gaussian distribution $p(\mathbf{x})$ with mean $\mu$ and a $d\times d$ positive definite covariance matrix $\Sigma$, 
its entropy is 
\begin{equation}
    H\left[p(\mathbf{x})\right]=\frac{m}{2}[1+\ln (2 \pi)]+\frac{\ln |\Sigma|}{2} .
\end{equation}
Therefore, the entropy of $p(\mathbf{z})$ and $q(\mathbf{x}_{t-1}|\mathbf{x}_{t},G_\phi\left(\mathbf{x}_t, \mathbf{z}, t\right))$ can be computed directly:
\begin{equation}
    H\left[p(\mathbf{z})\right]=\frac{d}{2}(1+\log (2 \pi))
\end{equation}
\begin{equation}
    H[q(\mathbf{x}_{t-1}|\mathbf{x}_{t},G_\phi\left(\mathbf{x}_t, \mathbf{z}, t\right))]=\frac{D}{2}(1+\log (2 \pi))+\frac{D}{2}\log\tilde{\beta_t}
\end{equation}

\subsection{Derivations of Eq.~\eqref{entropy upper bound}}
\begin{equation}
\begin{aligned}
     &H[p_\phi(\mathbf{z}|\mathbf{x}_{t-1},\mathbf{x}_{t})] 
     =-\mathbb{E}_{q(\mathbf{x}_t)p_\phi(\mathbf{x}_{t-1},\mathbf{z}|\mathbf{x}_{t})} \log p_\phi(\mathbf{z}|\mathbf{x}_{t-1},\mathbf{x}_{t})\\
    & =-\mathbb{E}_{q(\mathbf{x}_t)p_\phi(\mathbf{x}_{t-1},\mathbf{z}|\mathbf{x}_{t})} \log \frac{p_\phi(\mathbf{z}|\mathbf{x}_{t-1},\mathbf{x}_{t})}{q_\psi(\mathbf{z}|\mathbf{x}_{t-1},\mathbf{x}_{t})}-\mathbb{E}_{q(\mathbf{x}_t)p_\phi(\mathbf{x}_{t-1},\mathbf{z}|\mathbf{x}_{t})} \log q_\psi(\mathbf{z}|\mathbf{x}_{t-1},\mathbf{x}_{t})\\
    &=-\mathbb{E}_{q(\mathbf{x}_t)p_\phi(\mathbf{x}_{t-1}|\mathbf{x}_{t})}D_{\mathrm{KL}}(p_\phi(\mathbf{z}|\mathbf{x}_{t-1},\mathbf{x}_{t})\|q_\psi(\mathbf{z}|\mathbf{x}_{t-1},\mathbf{x}_{t}))-\mathbb{E}_{q(\mathbf{x}_t)p_\phi(\mathbf{x}_{t-1},\mathbf{z}|\mathbf{x}_{t})} \log q_\psi(\mathbf{z}|\mathbf{x}_{t-1},\mathbf{x}_{t})\\
    & \leq -\mathbb{E}_{q(\mathbf{x}_t)p_\phi(\mathbf{x}_{t-1},\mathbf{z}|\mathbf{x}_{t})} \log q_\psi(\mathbf{z}|\mathbf{x}_{t-1},\mathbf{x}_{t})
     \end{aligned}
\end{equation}

\subsection{Derivations of Eq.~\eqref{ELBO}}
\begin{equation}
	\begin{aligned}
		\log p_\phi(\mathbf{x}_{t-1}|\mathbf{x}_{t}) &= \log \int p(\mathbf{z}) q\left(\mathbf{x}_{t-1} \mid \mathbf{x}_t, \mathbf{x}_0=G_\phi\left(\mathbf{x}_t, \mathbf{z}, t\right)\right) \mathrm{d} \mathbf{z}\\
  &= \log \int q_\psi(\mathbf{z}|\mathbf{x}_{t-1},\mathbf{x}_{t})\frac{p(\mathbf{z})}{q_\psi(\mathbf{z}|\mathbf{x}_{t-1},\mathbf{x}_{t})}q\left(\mathbf{x}_{t-1} \mid \mathbf{x}_t, \mathbf{x}_0=G_\phi\left(\mathbf{x}_t, \mathbf{z}, t\right)\right) \mathrm{d} \mathbf{z}\\
  &\geq \int q_\psi(\mathbf{z}|\mathbf{x}_{t-1},\mathbf{x}_{t}) \left[\log \frac{p(\mathbf{z})}{q_\psi(\mathbf{z}|\mathbf{x}_{t-1},\mathbf{x}_{t})}+\log q\left(\mathbf{x}_{t-1} \mid \mathbf{x}_t, \mathbf{x}_0=G_\phi\left(\mathbf{x}_t, \mathbf{z}, t\right)\right)\right] \mathrm{d} \mathbf{z}\\
 & =
  -D_{K L}\left(q_\psi(\mathbf{z}|\mathbf{x}_{t-1},\mathbf{x}_{t} ) \| p(\mathbf{z})\right)+\mathbb{E}_{q_\psi(\mathbf{z}|\mathbf{x}_{t-1},\mathbf{x}_{t})}\log q(\mathbf{x}_{t-1}|\mathbf{x}_{t}, G_\phi\left(\mathbf{x}_t, \mathbf{z}, t\right))
	\end{aligned}
\end{equation}

\subsection{Derivations of Eq.~\eqref{second term}}
\begin{equation}
	\begin{aligned}
&\sum_{t \geq 1}\mathbb{E}_{q\left(\mathbf{x}_t\right)} D_{\mathrm{KL}}(p_\theta(\mathbf{x}_{t-1}|\mathbf{x}_{t})\|p_\phi(\mathbf{x}_{t-1}|\mathbf{x}_{t}))\\
=&\sum_{t \geq 1}\mathbb{E}_{q\left(\mathbf{x}_t\right)p_\theta(\mathbf{x}_{t-1}|\mathbf{x}_{t})}\log p_\theta(\mathbf{x}_{t-1}|\mathbf{x}_{t})-\mathbb{E}_{q\left(\mathbf{x}_t\right)p_\theta(\mathbf{x}_{t-1}|\mathbf{x}_{t})}\log p_\phi(\mathbf{x}_{t-1}|\mathbf{x}_{t})
			\end{aligned}
\end{equation}
The first term can be disregarded since it's independent of $\phi$, thus the variational upper bound of Eq.~\eqref{second} can be written as follows by applying Eq.~\eqref{ELBO} to the above equation:
\begin{equation}
\begin{aligned}
     -\sum_{t \geq 1}\mathbb{E}_{q\left(\mathbf{x}_t\right)p_\theta(\mathbf{x}_{t-1}|\mathbf{x}_{t})}\log p_\phi(\mathbf{x}_{t-1}|\mathbf{x}_{t})
    & \leq \sum_{t  \geq 1}\mathbb{E}_{q\left(\mathbf{x}_t\right)p_\theta(\mathbf{x}_{t-1}|\mathbf{x}_{t})}D_{K L}\left(q_\psi(\mathbf{z}|\mathbf{x}_{t-1},\mathbf{x}_{t} ) \| p(\mathbf{z})\right)\\
     &-\mathbb{E}_{q\left(\mathbf{x}_t\right)p_\theta(\mathbf{x}_{t-1}|\mathbf{x}_{t}) q_\psi(\mathbf{z}|\mathbf{x}_{t-1},\mathbf{x}_{t})}\log q(\mathbf{x}_{t-1}|\mathbf{x}_{t}, G_\phi\left(\mathbf{x}_t, \mathbf{z}, t\right))
\end{aligned}
\end{equation}
\section{Experimental details}
\subsection{Diffusion process}
We use the discretization of the continuous-time VP SDE~\cite{song2020score} as our diffusion process, which is the same as DDGAN. For all datasets, we set the number of time steps $T$ to be 4. The variance function of VP SDE is given by:
\begin{equation}
	\sigma^2\left(t^{\prime}(t)\right)=1-e^{-\beta_{\min } t^{\prime}(t)-0.5\left(\beta_{\max }-\beta_{\min }\right) t^{\prime 2}(t)},
	\label{variance_schedule}
\end{equation}
where $t^{\prime}(t)\in [0,1]$, $t^{\prime}(t)$ is a function of time step $t$ denoting the time of $\mathbf{x}_t$ in VP SDE. $t^{\prime}(t)$ can be any flexible time schedule of a forward SDE. The constants $\beta_{\max}$ and $\beta_{\min}$ are chosen differently for different datasets. $\beta_t$ is designed as follows:
\begin{equation}
	\beta_t=1-\frac{1-\sigma^2\left(t^{\prime}(t)\right)}{1-\sigma^2\left(t^{\prime}(t-1)\right)}=1-e^{-\beta_{\min }\left(t^{\prime}(t)-t^{\prime}(t-1)\right)-0.5\left(\beta_{\max }-\beta_{\min }\right)\left(t^{\prime 2}(t)-t^{\prime 2}(t-1)\right)},
\end{equation}
where $t \in\{0, 1,2, \ldots T\}$. Except for LSUN church dataset, we use equidistant steps in time for $t^{\prime}(t)$ on other datasets, i.e. $t^{\prime}(t)=\frac{t}{T}$. For LSUN church, we borrow the time schedule from DRL, which focuses on the first stage of the diffusion process:
$t^{\prime}(t)=\left\{\begin{array}{ll}
\frac{t\lfloor\frac{500}{T-1}\rfloor}{999} & t\le T \\
1 & t =T
\end{array} \right.$, we assume $T\leq 999$ in most cases. We choose this time schedule because 
signal-to-noise ratio (SNR)~\cite{kingma2021variational} is strictly monotonically decreasing in time, in high-dimensional space, data becomes quite sparse, thus training more on the first stage may be more important for generation. 
\subsection{Network structure}
\begin{table*}[thbp]
	\centering
	\caption{Network structures for toy datasets. BN denotes batch normalization. $SE$ denotes sinusoidal embedding. $\mathbf{x}_{out},t_{out}$ denote the outputs of two embedding networks with $\mathbf{x}_t$-related input and sinusoidal embedding of $t$ as the inputs.}
	\label{network_toy}
	\begin{minipage}{0.33\textwidth}
		\centering
		\begin{tabular}{clcl}
			\hline
			\multicolumn{4}{c}{Energy}                                                                                                                                                                             \\ \hline
			\multicolumn{2}{c|}{$\mathbf{x}_t$}                                                                          & \multicolumn{2}{c}{$SE(t)$}                                                                             \\ \hline
			\multicolumn{2}{c|}{\multirow{2}{*}{\begin{tabular}[c]{@{}c@{}}FC 16 PReLU\\  FC 32\end{tabular}}} & \multicolumn{2}{c}{\multirow{2}{*}{\begin{tabular}[c]{@{}c@{}}FC 16 PReLU\\  FC 32\end{tabular}}} \\
			\multicolumn{2}{c|}{}                                                                              & \multicolumn{2}{c}{}                                                                              \\ \hline
			\multicolumn{4}{c}{\begin{tabular}[c]{@{}c@{}}concat {[}$\mathbf{x}_{out},t_{out}${]}\\ FC 300 PReLU\\ FC 300 PReLU\\ FC 1\end{tabular}}                                                                               \\ \hline
		\end{tabular}%
	\end{minipage}%
	\begin{minipage}{0.33\textwidth}
		\centering
		\begin{tabular}{clcl}
			\hline
			\multicolumn{4}{c}{Generator}                                                                                                                                                                             \\ \hline
			\multicolumn{2}{c|}{concat $[\mathbf{x}_t$,$\mathbf{z}]$}                                                             & \multicolumn{2}{c}{$SE(t)$}                                                                             \\ \hline
			\multicolumn{2}{c|}{\multirow{2}{*}{\begin{tabular}[c]{@{}c@{}}FC 16 PReLU BN\\  FC 32\end{tabular}}} & \multicolumn{2}{c}{\multirow{2}{*}{\begin{tabular}[c]{@{}c@{}}FC 16 PReLU\\  FC 32\end{tabular}}} \\
			\multicolumn{2}{c|}{}                                                                                 & \multicolumn{2}{c}{}                                                                              \\ \hline
			\multicolumn{4}{c}{\begin{tabular}[c]{@{}c@{}}concat $[\mathbf{x}_{out},t_{out}]$\\ FC 300 PReLU BN\\ FC 300 PReLU BN\\ FC 2\end{tabular}}                                                                             \\ \hline
		\end{tabular}%
	\end{minipage}%
	\begin{minipage}{0.33\textwidth}
		\centering
		\begin{tabular}{clcl}
			\hline
			\multicolumn{4}{c}{Encoder}                                                                                                                                                                               \\ \hline
			\multicolumn{2}{l|}{concat $[\mathbf{x}_{t-1},\mathbf{x}_t]$}                                                         & \multicolumn{2}{c}{$SE(t)$}                                                                             \\ \hline
			\multicolumn{2}{c|}{\multirow{2}{*}{\begin{tabular}[c]{@{}c@{}}FC 16 PReLU BN\\  FC 32\end{tabular}}} & \multicolumn{2}{c}{\multirow{2}{*}{\begin{tabular}[c]{@{}c@{}}FC 16 PReLU\\  FC 32\end{tabular}}} \\
			\multicolumn{2}{c|}{}                                                                                 & \multicolumn{2}{c}{}                                                                              \\ \hline
			\multicolumn{4}{c}{\begin{tabular}[c]{@{}c@{}}concat $[\mathbf{x}_{out},t_{out}]$\\ FC 300 PReLU BN\\ FC 300 PReLU BN\\ FC 16\end{tabular}}                                                                            \\ \hline
			\multicolumn{2}{c}{mean: FC 2}                                                                        & \multicolumn{2}{c}{logvar: FC 2}                                                                  \\ \hline
		\end{tabular}%
	\end{minipage}
\end{table*}
\begin{table*}[thbp]
	\centering
	\caption{Network structures for MNIST dataset. $permutate(\cdot)$ denotes the permutation operation in Glow.}
	\label{network_mnist}
	\begin{minipage}{0.33\textwidth}
		\centering
		\begin{tabular}{ccl}
			\hline
			\multicolumn{3}{c}{Energy}                                      \\ \hline
			\multicolumn{1}{c|}{$permutate(\mathbf{x}_t)$}     & \multicolumn{2}{c}{$SE(t)$}     \\ \hline
			\multicolumn{3}{c}{4 NICE layers}                               \\
			\multicolumn{3}{c}{NICE scale layer}                            \\ \hline \vspace{0.01pt}
		\end{tabular}
		\vspace{0.1pt}
		\begin{tabular}{ccl}
			\hline
			\multicolumn{3}{c}{NICE layer}                                  \\ \hline
			\multicolumn{2}{c}{FC 1000 PReLU}         & \multirow{2}{*}{$\times 4$} \\
			\multicolumn{2}{c}{+ FC 1000 PReLU~($SE(t)$)} &                     \\\hline
			\multicolumn{3}{c}{FC 392}                                      \\ \hline
		\end{tabular}%
	\end{minipage}%
	\begin{minipage}{0.33\textwidth}
		\centering
		\begin{tabular}{clcl}
			\hline
			\multicolumn{4}{c}{Generator}                                                                                                                                                                             \\ \hline
			\multicolumn{2}{c|}{concat $[\mathbf{x}_t$,$\mathbf{z}]$}                                                             & \multicolumn{2}{c}{$SE(t)$}                                                                             \\ \hline
			\multicolumn{2}{c|}{\multirow{2}{*}{\begin{tabular}[c]{@{}c@{}}FC 1600 PReLU \\  FC 1600\end{tabular}}} & \multicolumn{2}{c}{\multirow{2}{*}{\begin{tabular}[c]{@{}c@{}}FC 50 PReLU\\  FC 100\end{tabular}}} \\
			\multicolumn{2}{c|}{}                                                                                 & \multicolumn{2}{c}{}                                                                              \\ \hline
			\multicolumn{4}{c}{\begin{tabular}[c]{@{}c@{}}concat $[\mathbf{x}_{out},t_{out}]$\\ FC 3200 PReLU BN\\ FC 1600 PReLU BN\\ FC 784\end{tabular}}                                                                             \\ \hline
		\end{tabular}%
	\end{minipage}%
	\begin{minipage}{0.33\textwidth}
		\centering
		\begin{tabular}{clcl}
			\hline
			\multicolumn{4}{c}{Encoder}                                                                                                                                                                               \\ \hline
			\multicolumn{2}{l|}{concat $[\mathbf{x}_{t-1},\mathbf{x}_t]$}                                                         & \multicolumn{2}{c}{$SE(t)$}                                                                             \\ \hline
			\multicolumn{2}{c|}{\multirow{2}{*}{\begin{tabular}[c]{@{}c@{}}FC 3200 PReLU\\  FC 1600\end{tabular}}} & \multicolumn{2}{c}{\multirow{2}{*}{\begin{tabular}[c]{@{}c@{}}FC 50 PReLU\\  FC 100\end{tabular}}} \\
			\multicolumn{2}{c|}{}                                                                                 & \multicolumn{2}{c}{}                                                                              \\ \hline
			\multicolumn{4}{c}{\begin{tabular}[c]{@{}c@{}}concat $[\mathbf{x}_{out},t_{out}]$\\ FC 1600 PReLU BN\\ FC 1600 PReLU BN\end{tabular}}                                                                            \\ \hline
			\multicolumn{2}{c}{mean: FC 50}                                                                        & \multicolumn{2}{c}{logvar: FC 50}                                                                  \\ \hline
		\end{tabular}%
	\end{minipage}
\end{table*}
\subsubsection{Toy data}
For toy datasets, our encoder, generator, and energy function all have two embedding networks to encode the toy data, latent variable, and time $t$ into two features. Then a decoder is used to decode the concatenation of these features into our desired output. Network structures are illustrated in Table~\ref{network_toy}.  We use the Adam optimizer with a learning rate of $10^{-4}$ for all the networks. The batch size is 200, and we train the model for 180k iterations. For FCE, we also choose Glow as the generator, where we use 5 affine coupling layers amounting to 15 fully connected layers with 300 hidden units. For other adversarial EBMs, both the generator and energy function have 3 fully connected layers each with 300 hidden units and PReLU activations. Batch normalization is used in generators. The latent dimension is set to 2.

\subsubsection{MNIST}
For our NICE\_t network, similar to NICE model in VERA~\cite{nomcmc}, we also have 4 coupling layers and each coupling layer has 5 hidden layers with 1000 units and PReLU nonlinearity. We integrate parameter $t$ into NICE by adding a linear layer with the time embeddings as the inputs and PReLU nonlinearity to the output of each hidden layer. For other WGAN-based methods, we use the same NICE model as in VERA. We deepen their generators, which allows for a fairer comparison of our method. Their generators all have a latent dimension of 100 and 5 hidden layers with 1600 or 3200 units each. We adopt PReLU nonlinearity and batch normalization in their generators as is common with generator networks. Our generator and encoder employ the same structure as those for toy datasets except for more hidden units. The latent dimension is set to 50. Network structures are shown in Table~\ref{network_mnist}. All models were trained for 400 epochs with the Adam optimizer. We use the learning rate $10^{-4}$ for all our networks. For other WGAN-based methods, we use learning rate $3\times 10^{-6}$ for energy function and $3\times 10^{-4}$ for generator. We use a batch size of 128 for all models.

\subsubsection{Large-scale datasets}
For large-scale datasets, we use sinusoidal positional embeddings for conditioning on integer time steps. For CIFAR-10 and CelebA datasets, our generator follows the modified Unet structure in DDGAN~\cite{xiao2021tackling}, which provides z-conditioning to the NCSN++ architecture~\cite{song2020score}. Our energy function mostly follows NCSN++ in Score SDE, which also takes $\mathbf{x}_t$ and $t$ as its inputs. The only difference is we remove the last scale-by-sigma operation in NCSN++ and replace it with a negative $\ell_2$ norm between input $\boldsymbol{x}_{t}$ and the output of the NCSN++ network as in Eq.~\eqref{energy_definition}. Our encoder is designed as a CNN network that incorporates time embeddings and concatenates $\mathbf{x}_{t-1}$ and $\mathbf{x}_t$. The Encoder network is shown in Table~\ref{network_encoder}. 
For LSUN dataset, we simply removed FIR upsampling/downsampling and progressive growing architecture from the z-conditioned NCSN++ and named it DDPM++ as our generator, following the naming convention of Score SDE. For energy function, we adopt DDPM++ in Score SDE except for the same modifications used in NCSN++. We find that DDPM++ helps generation on LSUN dataset. Our encoder borrows the network structure of the discriminator in DDGAN, but replaces the final dense layer with two dense layers that separately output the mean and variance of $q_\psi(\mathbf{z}|\mathbf{x}_{t-1},\mathbf{x}_{t})$. See DDGAN for more details.

\begin{table*}[htbp]
	\centering
	\caption{Encoder structures for CIFAR-10 and CelebA datasets.}
	\label{network_encoder}
 \vspace{5pt}
	\begin{minipage}[t]{0.45\textwidth}
	\centering
		\begin{tabular}{cc}
			\hline
			\multicolumn{2}{c}{Encoder of CIFAR-10}                                                                                                                                                                                                                                        \\ \hline
			\multicolumn{1}{c|}{concat $[\mathbf{x}_{t-1},\mathbf{x}_t]$}                                                            & \begin{tabular}[c]{@{}c@{}}$SE(t)$\\ FC 256 LeakyReLU~(0.2)\\ FC 256 LeakyReLU~(0.2)\end{tabular}                                                            \\ \hline
			\multicolumn{2}{c}{\begin{tabular}[c]{@{}c@{}}3$\times$3 Conv2D 64, LeakyReLU~(0.2)\\ + Dense 64~($t_{out}$)\\ 4$\times$4 Conv2D 128, LeakyReLU~(0.2)\\ +Dense 128($t_{out}$)\\ 4$\times$4 Conv2D 256, LeakyReLU~(0.2)\\ +Dense 256~($t_{out}$)\\ 4*4 Conv2D  512, LeakyReLU~(0.2)\\ +Dense 512~($t_{out}$)\\ \end{tabular}} \\ \hline
			\multicolumn{1}{l|}{mean: 4$\times$4 Conv2D 100}                                                               & logvar: 4$\times$4 Conv2D 100                                                                                                                                   \\ \hline
		\end{tabular}%
	\end{minipage}
	\hfill
	\begin{minipage}[t]{0.45\textwidth}
		\centering
		\begin{tabular}{cc}
		\hline
		\multicolumn{2}{c}{Encoder of CelebA}                                                                                                                                                                                                                                        \\ \hline
		\multicolumn{1}{c|}{concat $[\mathbf{x}_{t-1},\mathbf{x}_t]$}                                                            & \begin{tabular}[c]{@{}c@{}}$SE(t)$\\ FC 256 LeakyReLU~(0.2)\\ FC 256 LeakyReLU~(0.2)\end{tabular}                                                            \\ \hline
		\multicolumn{2}{c}{\begin{tabular}[c]{@{}c@{}}3$\times$3 Conv2D 64, LeakyReLU~(0.2)\\ + Dense 64~($t_{out}$)\\ 4$\times$4 Conv2D 128, LeakyReLU~(0.2)\\ +Dense 128($t_{out}$)\\ 4$\times$4 Conv2D 256, LeakyReLU~(0.2)\\ +Dense 256~($t_{out}$)\\ 4*4 Conv2D  512, LeakyReLU~(0.2)\\ +Dense 512~($t_{out}$)\\ 4$\times$4 Conv2D 1024, LeakyReLU~(0.2) \\ +Dense 1024 ($t_{out}$)\end{tabular}} \\ \hline
		\multicolumn{1}{l|}{mean: 4$\times$4 Conv2D 100}                                                               & logvar: 4$\times$4 Conv2D 100                                                                                                                                   \\ \hline
	\end{tabular}%
	\end{minipage}
\end{table*}
\subsection{Hyperparameter settings}
We specify the hyperparameters used for our generators and training optimization on each dataset in Table~\ref{hyperparam_generator} and Table~\ref{hyperparam_optim}.
\begin{table}[htbp]
	\centering
	\caption{Hyper-parameters for our generator network}
	\label{hyperparam_generator}
	\resizebox{0.7\linewidth}{!}{%
		\begin{tabular}{cccc}
			\hline
			& CIFAR-10  & CelebA  & LSUN church    \\ \hline
			\# of ResNet blocks per scale     & 2         & 2    & 2       \\
			Initial \# of channels            & 128       & 64     & 64     \\
			Channel multiplier for each scale & (1,2,2,2) & (1,1,2,2,4) & (1,2,2,4,4) \\
			Scale of attention block          & 16        & 16    & 16       \\
			Latent Dimension                  & 100       & 100   & 100       \\
			\# of latent mapping layers       & 4         & 4      & 4        \\
			Latent embedding dimension        & 256       & 256    & 256     \\ \hline
		\end{tabular}%
	}
\end{table}
\begin{table}[htbp]
	\centering
	\caption{Hyper-parameters for our training optimization}
	\label{hyperparam_optim}
	\resizebox{0.7\linewidth}{!}{%
		\begin{tabular}{ccccc}
			\hline
			& MNIST      & CIFAR-10   & CelebA  & LSUN church   \\ \hline
			Initial learning rate            & 1e-4       & 1e-4       & 5e-5  & 5e-5     \\
			$\beta_{min}, \beta_{max}$ in Eq.~(\ref{variance_schedule}) & (0.1, 10) & (0.1, 20) & (0.1, 20) & (0.1, 20) \\
            $w$, $w_{\text {mid }}$ in Eq.~(\ref{weight elbo}) & (1, 1) & (1, 1) & (1, 1) & (0.6, 0.2) \\
			Adam optimizer $\beta_1,\beta_2$ & (0.0, 0.9) & (0.0, 0.9) & (0.0, 0.9) & (0.0, 0.9) \\
			EMA                              & None       & 0.9999     & 0.999  & 0.999    \\
			Batch size                       & 128        & 64         & 32     & 12     \\
			\# of training epochs            & 400        & 1200       & 400  & 400      \\
			\# of GPUs                       & 1          & 4          & 4    & 4      \\ \hline
		\end{tabular}%
	}
\end{table}
\subsection{Evaluation}
When evaluating FID and IS scores, we use 50k generated samples for CIFAR-10, CelebA, and LSUN church datasets. We use Pytorch 1.10.0 and CUDA 11.3 for training. Our training converges approximately 3 times faster than DDGAN, resulting in a comparable overall training time despite our model's slower per-epoch training speed compared to DDGAN.
\section{Additional results}
\subsection{More comparisons on Toy dataset}
\begin{figure}[htbp]
		\footnotesize
		\centering
		\renewcommand{\tabcolsep}{1pt} \renewcommand{\arraystretch}{0.1}
  \resizebox{0.8\linewidth}{!}{%
		\begin{tabular}{cccccc}
				\includegraphics[width=0.14\linewidth]{real_pinwheel2.png} &
				\includegraphics[width=0.14\linewidth]{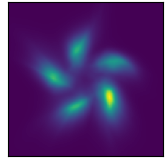} &
				\includegraphics[width=0.14\linewidth]{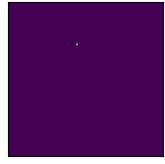} &
				\includegraphics[width=0.14\linewidth]{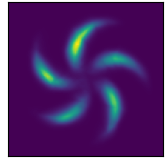}&
				\includegraphics[width=0.14\linewidth]{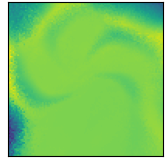}&
				\includegraphics[width=0.14\linewidth]{pr_ddEBM_pinwheel.png} \\
			real data & Score flow & \begin{tabular}[c]{@{}c@{}}DRL\\  w/o scale\end{tabular} & \begin{tabular}[c]{@{}c@{}}DRL\\ w/ scale 0.01\end{tabular} & DDGAN & DDAEBM
			\end{tabular}}
		\caption{Density estimation on pinwheel dataset for different methods.}
		\label{density_estimation_compare} 
	\end{figure} 
 \begin{figure}[htbp]
		\footnotesize
		\centering
		\renewcommand{\tabcolsep}{1pt} \renewcommand{\arraystretch}{0.1}
    \resizebox{0.8\linewidth}{!}{%
		\begin{tabular}{cccccc}
				\includegraphics[width=0.14\linewidth]{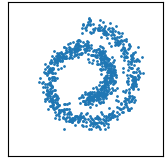} &
				\includegraphics[width=0.14\linewidth]{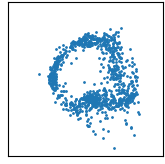} &
				\includegraphics[width=0.14\linewidth]{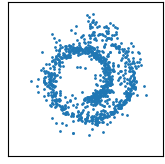} &
				\includegraphics[width=0.14\linewidth]{real_25gaussians.png}&
				\includegraphics[width=0.14\linewidth]{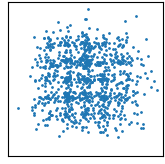}&
				\includegraphics[width=0.14\linewidth]{gene3_ddEBM_25gaussians.png} \\
			Swiss roll& DDGAN  & DDAEBM &  Swiss roll & DDAEBM~(KL) & DDAEBM~(Jeffrey)
			\end{tabular}}
		\caption{More generation results on Swiss roll and 25-Gaussians datasets}
		\label{more generation} 
	\end{figure} 
Fig~\ref{density_estimation_compare} further demonstrates our model's strong capability on density estimation by comparing with score flow~\cite{song2021maximum}, DRL and DDGAN. Score flow and DRL specialize in density estimation, especially for large-scale datasets, but they can't perform well on toy datasets. 
We choose sub-VP SDE for the diffusion process and replace the score network with the gradient of an energy function, the network of which is the same as ours except for changing positional embeddings to random Fourier feature embeddings for conditioning on continuous time steps.                                                                                              
DRL requires multiplying its energy function by a scaling factor of 0.01, which introduces a temperature parameter between the trained energy function and the target one.
DDGAN combines the diffusion process with GAN instead of EBM, as we expected, DDGAN is not suitable as a density estimator because GAN-based models are not supposed to provide a density estimate.

We also show more generation results in Fig.~\ref{more generation} to compare our model with DDGAN and DDAEBM trained with KL divergence. Although DDGAN can achieve impressive sample quality on the 25-Gaussians dataset, as demonstrated in its initial paper, it gets limited quality on the Swiss roll dataset, which means DDGAN is not stable and robust enough across various datasets. Our DDAEBM can get much better generation performance on Swiss roll dataset, verifying our model's advantage in terms of robustness.
We can also observe that without the reversed KL divergence, DDAEBM fails to converge to each mode, leading to poor sample quality. Our DDAEBM trained with symmetric Jeffrey divergence can improve generation a lot, resulting in satisfactory sample quality on each mode. This experiment demonstrates besides the training of energy function as shown in Table~\ref{ablation_modif},
training with Jeffrey divergence also has superiority to the generator's training by adding an extra reversed KL divergence.
 \begin{table}[htbp]
	\centering
	\caption{Mode coverage on StackedMNIST}
	\label{Mode coverage}
	\begin{tabular}{clcc}
		\hline
		\multicolumn{2}{c}{Model}        & Modes         & KL             \\ \hline
		\multicolumn{2}{c}{VEEGAN~\cite{Srivastava_Valkov_Russell_Gutmann_Sutton_2017}}       & 762           & 2.173          \\
		\multicolumn{2}{c}{PresGAN~\cite{dieng2019prescribed}}      & \textbf{1000} & 0.115          \\
		\multicolumn{2}{c}{StyleGAN2~\cite{karras2020analyzing}}    & 940           & 0.424          \\
		\multicolumn{2}{c}{Adv.DSM~\cite{jolicoeur2020adversarial}}      & \textbf{1000} & 1.49           \\
		\multicolumn{2}{c}{VAEBM~\cite{xiao2020vaebm}}        & \textbf{1000} & 0.087          \\
		\multicolumn{2}{c}{DDGAN~\cite{xiao2021tackling}}        & \textbf{1000} & 0.071          \\
		\multicolumn{2}{c}{MEG~\cite{kumar2019maximum}}        & \textbf{1000} & 0.042          \\
		\multicolumn{2}{c}{EBM-BB~\cite{geng2021bounds}}        & \textbf{1000} & 0.045          \\
		 \hline
		\multicolumn{2}{c}{DDAEBM(ours)}  & \textbf{1000} & \textbf{0.033} \\ \hline
	\end{tabular}
\end{table}
 \subsection{Mode counting}
Generative Adversarial Networks~(GANs) are notorious for mode collapse which is a phenomenon in which the generator function maps all samples to a small subset of the observation space. It's well known that EBM can alleviate mode collapse because of its entropy term. Therefore, we also evaluate the mode coverage of our model on the StackedMNIST dataset. StackedMNIST is a synthetic dataset that contains images generated by randomly choosing three MNIST images and stacking them along the RGB channels. Hence the true total number of modes is 1,000, and they are counted using a pretrained MNIST classifier. Similar to DDGAN, we also follow the setting of~\cite{Lin_Khetan_Fanti_Oh_2020} and report the number of covered modes and the KL divergence from the categorical distribution over 1000 categories of generated samples to true data in Table~\ref{Mode coverage}. It shows that our model covers all modes and achieves the lowest KL compared to GAN-based models and EBMs. This demonstrates our model is effective in mitigating mode collapse.
\subsection{Additional results on LSUN church}
\begin{figure}[htbp]
	\footnotesize
	\centering
	\includegraphics[width=0.9\linewidth]{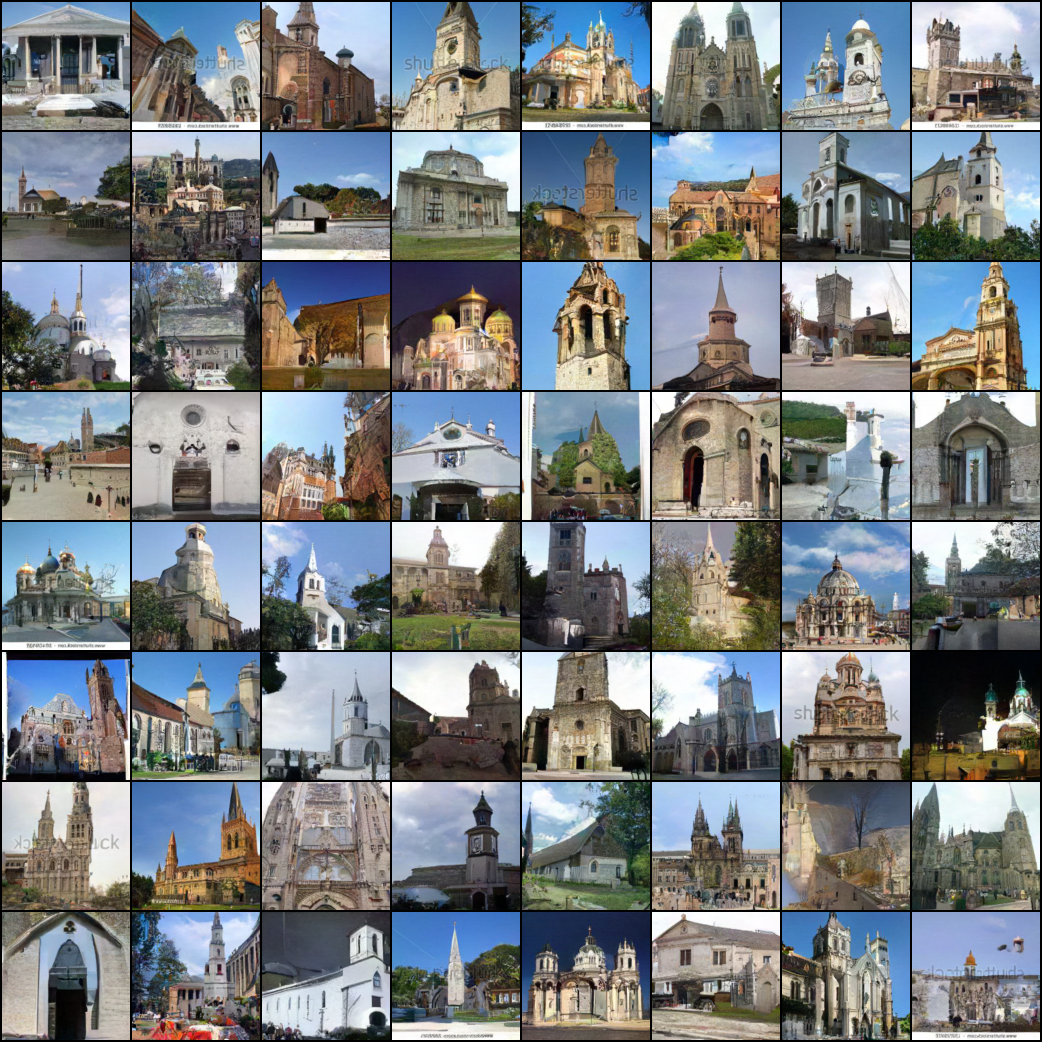} 		
	\caption{Generated samples using DDAEBM on LSUN church dataset. FID=13.80}
	\label{lsun_ebm} 
\end{figure}

\begin{figure}[htbp]
	\footnotesize
	\centering
	\includegraphics[width=0.9\linewidth]{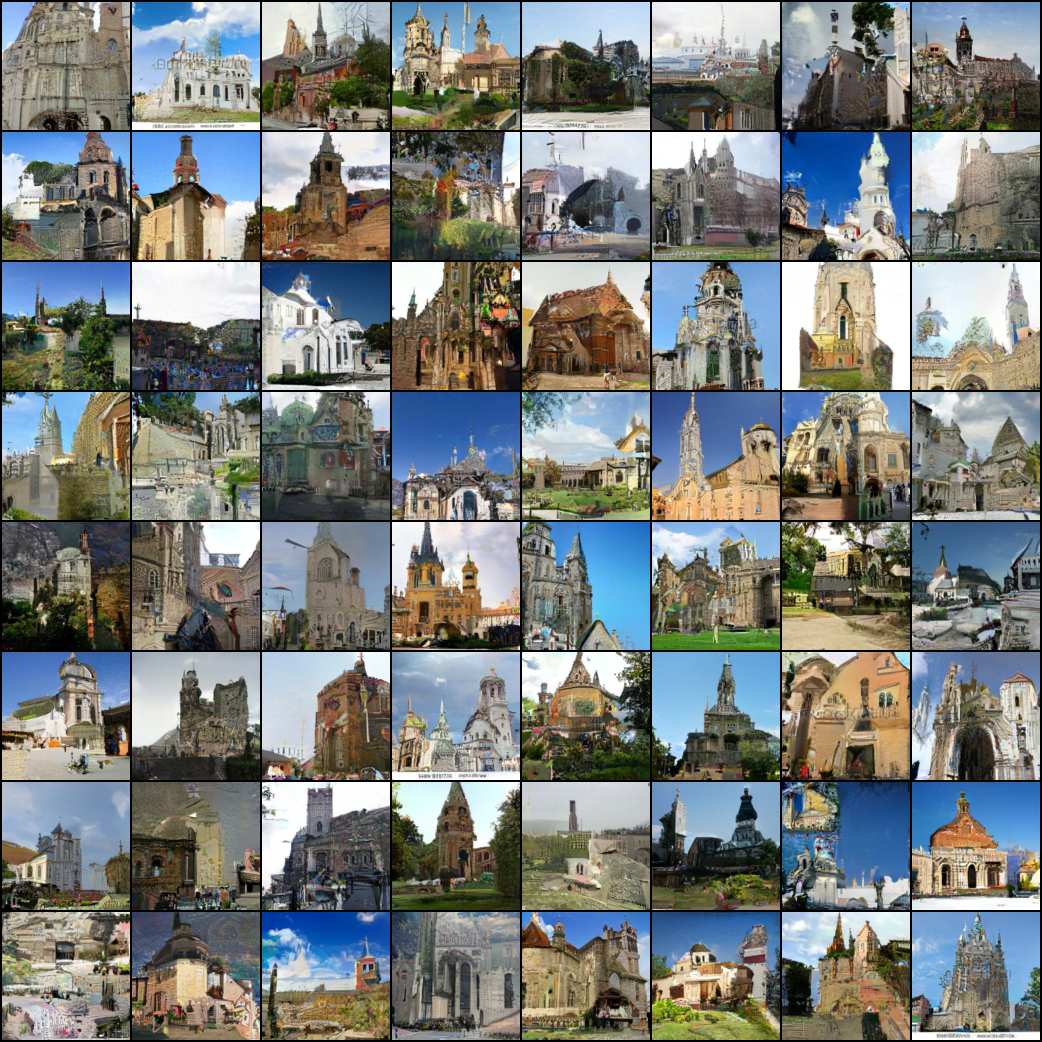} 		
	\caption{Generated samples using DRL on LSUN church dataset. FID=26.69}
	\label{lsun_DRL} 
\end{figure}

We provide more results on $128\times 128$ LSUN church dataset. Since there are few baselines on this resolution, we just compare our model with DRL~\cite{gao2020learning}, which is superior to the majority of EBMs on various image datasets. Our DDAEBM can get better visual quality compared to DRL as illustrated in Fig.~\ref{lsun_ebm} and Fig.~\ref{lsun_DRL}. We calculate FID on 50,000 samples using a PyTorch implementation from DDGAN. Our DDAEBM has 13.80 FID. Original paper of DRL reports 9.76 FID with TensorFlow for calculation. But we used officially provided checkpoints and evaluated FID using our PyTorch implementation, we only got 26.69 FID which is much worse than ours.

\end{document}